\documentclass[10pt,twocolumn,letterpaper]{article}
\usepackage{cvpr}

\definecolor{cvprblue}{rgb}{0.21,0.49,0.74}
\usepackage{tikz}
\usetikzlibrary{positioning,arrows.meta,fit}
\usepackage[breaklinks,colorlinks,allcolors=cvprblue]{hyperref}
\usepackage{flafter}
\newcommand{\fitwidth}[1]{%
  \sbox0{#1}%
  \ifdim\wd0>\columnwidth
    \resizebox{\columnwidth}{!}{\usebox0}%
  \else
    \usebox0
  \fi}

\title{Pictura: Generating Images with Strokes}
\hypersetup{pdftitle={Pictura: Generating Images with Strokes}}
\author{Ze-Xuan Zheng, Xi-Zhen Zhang, Jun-Jie Luo, Guoyuan Lu, Zhaokun Lin, Bao Qin%
\thanks{Bao Qin is the corresponding author.}}

\makeatletter
\let\picturaorig@maketitle\@maketitle
\def\@maketitle{\picturaorig@maketitle
  \vspace{-0.5em}
  \begin{center}
    \includegraphics[width=0.30\textwidth]{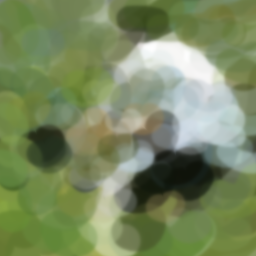}
  \end{center}
  \vspace{-0.6em}
}
\makeatother

\begin{document}
\maketitle
\pagestyle{plain}

\begin{abstract}
\thispagestyle{plain}
A person draws a scene in strokes; a text-to-image model paints it in pixels. We
ask whether images can be generated the way people draw them, by placing a few structured strokes rather than pixel by pixel. We build Pictura: a small DiT that
maps a noise canvas and a frozen CLIP embedding to $256$ B\'{e}zier strokes, which
a differentiable renderer composites into an image in one forward pass. Its output
is $3{,}584$ stroke parameters rather than $196{,}608$ pixel values, and editing it
means moving, reshaping or recolouring a stroke, not repainting pixels. On the
$1{,}000$ COCO images it was trained on, Pictura reaches $19.13$\,dB PSNR against
$12.01$\,dB for a standard diffusion model of comparable size and budget. A
one-step pixel regression scores higher still on PSNR, but its output is a
deterministic function of the caption: it recalls images rather than generating
them. On a corpus of Stable Diffusion renders, the stroke-parameter space yields
the lowest KID: $0.264$, against $0.308$ in pixel space and $0.420$ in
ResNet-feature space. Three things come out of this: a controlled comparison that isolates the output representation, a decomposition that separates dimension from structure, and a one-step generator whose low-dimensional output can be edited directly. Whether the same holds at larger scale, and whether strokes can carry object-level meaning, remains open.
\end{abstract}

\section{Introduction}
\label{sec:intro}

People have drawn scenes in strokes for far longer than they have described them in pixels. Text-to-image generation, by contrast, is data-hungry: the dominant formulation asks a model to emit a dense, high-dimensional output: a raster of pixels, or a latent field
that decodes back to one. Because every spatial location is an independent
degree of freedom, learning that distribution requires both large amounts of
data and multi-step sampling. Yet a person describing a scene does not do so pixel by pixel. A draftsman lays down a sequence of strokes, each an intentional unit with a place, a direction, a length, a curvature, a thickness, and a color. A scene that would take $196{,}608$ numbers often takes a few hundred such strokes. This is efficient in the two ways that matter for learning. Sparsity:
the output is a few thousand parameters rather than a few hundred thousand.
Structure: a stroke parameter has a bounded role, and its valid settings form a
thin manifold of plausible strokes, whereas a pixel has no such role and the set of
plausible images is combinatorially large. Both should shrink the effective learning problem, and therefore the data a model needs to bind the text condition.

We instantiate this as a one-step generator: a noise tensor maps through a
transformer decoder to $256$ B\'{e}zier strokes ($3{,}584$ dimensions), which a
differentiable renderer composites into a $256\times256$ image. The structural
prior lives in the generator, so the model moves from noise to the stroke manifold
in a single forward pass; we compare it against pixel formulations of comparable
size under the same data, parameters, and frozen text encoder.

Trained on $1{,}000$ COCO images, the stroke generator reaches $19.13$\,dB PSNR in one forward pass,
while at full resolution the pixel baseline stays below the constant-mean
baseline of $12.51$\,dB. The scope is deliberately narrow: with $1{,}000$ images we do not attempt open-vocabulary generalization, which at this scale no model in our comparison acquires. We ask whether a model, given a fixed small corpus, can
reproduce the images its captions describe. Our contributions are:

\begin{itemize}
  \item A controlled comparison within one generative family: both arms start
  from noise and must produce an image. With the same data, parameter budget,
  text encoder, and training length, decoding the noise into $256$ semantic
  strokes in one pass reaches $19.13$\,dB, while $50$-step pixel diffusion stays
  below the constant-mean baseline ($12.01$\,dB). Within this regime, what the
  generator maps the noise onto dominates whether the text condition is bound.
  \item A decomposition of the two factors. Dimensionality matters: a
  $35\times35$ pixel output ($3{,}675$ dims) recovers above the baseline, so part
  of the gap is a high-dimension effect. At that same low dimension, however, the
  structured stroke output still leads by $2.7$\,dB, so dimension and structure
  act as separable factors in every cell we can reach, and the stroke
  representation satisfies both in a single step.
  \item A one-step generator system: noise maps to $256$ B\'{e}zier strokes
  ($3{,}584$ dimensions) in one forward pass and runs in tens of milliseconds. The
  output is interpretable at the parameter level: each stroke has a bounded
  geometric role (position, length, curvature, width, colour, opacity), and the
  set can be edited after generation. The structural prior sits in the generator, not in a multi-step score procedure.
\end{itemize}

\paragraph{Positioning: a substrate, not a benchmark entry.}
The claim is about representation, not about a PSNR record. A generator whose output is a set of parameterized strokes, with the renderer inside the model, trains at $1{,}000$ images and binds its text condition where pixel output at the same budget does not. It also exposes its structure instead of hiding it in a raster. What changes is the premise that pixel output is the only place a text-to-image model can live. Every component here is minimal (one renderer, one decoder, one corpus), so these numbers are a floor rather than a verdict.

\section{Related Work}
\label{sec:related}

\paragraph{Text-to-image generation, diffusion, and one-step models.}
Modern text-to-image systems, including DALL$\cdot$E, Imagen, and the Stable Diffusion
family~\cite{ramesh2022hierarchical,radford2021learning,rombach2022high}, couple
a frozen or jointly trained text encoder with a diffusion denoiser, and DiT
\cite{peebles2023scalable} has established the transformer as its standard
backbone. These systems achieve strong results at billions of parameters and
image--text pairs, and treat the output representation (pixels or their latent field) as a fixed premise, never as a variable. Studies of the same family
under limited data focus on overfitting and memorization~\cite{kim2023diffusion,carlini2023extracting};
the output representation is left unvaried. A separate line replaces iterative
sampling with a single forward pass (GANs, consistency and distillation methods~\cite{song2023consistency,salimans2022progressive,kim2024ctm}, and flow
matching~\cite{lipman2023flow}), and implicit neural
representations~\cite{sitzmann2020implicit} likewise change what a model predicts, not how it is trained. Across this work the output representation remains an unisolated design choice;
prior comparisons also change scale, recipe or task, whereas we vary only it.

\paragraph{Stroke and vector synthesis.}
Learning to Paint~\cite{huang2019learning} places strokes in sequence to
reconstruct a target image, and much later work follows the same
image-to-stroke pattern: CLIPasso~\cite{vinker2022clipasso} fits strokes to a
single image by per-image optimization, while
DiffSketcher~\cite{wang2023diffsketcher}, SVGDreamer~\cite{xing2024svgdreamer}, and
VectorFusion~\cite{jain2023vectorfusion} steer rasterized vectors with diffusion
or optimization on differentiable rasterizers~\cite{li2020diffvg}. A closely related line predicts
strokes in a single feed-forward pass: Paint
Transformer~\cite{liu2021paint}, Stylized Neural Painting~\cite{zou2021stylized},
StarVector~\cite{rodriguez2025starvector}, DeepSVG~\cite{carlier2020deepsvg}, and
Im2Vec~\cite{reddy2021im2vec} generate vector graphics from an image or from text.
These are the closest systems to ours and establish that a feed-forward stroke
generator is trainable; none compares it against a pixel-output model under a
shared data and parameter budget, which is the question we ask.

\section{Method}
\label{sec:method}

\begin{figure*}[tb]
\centering
\includegraphics[width=0.62\textwidth]{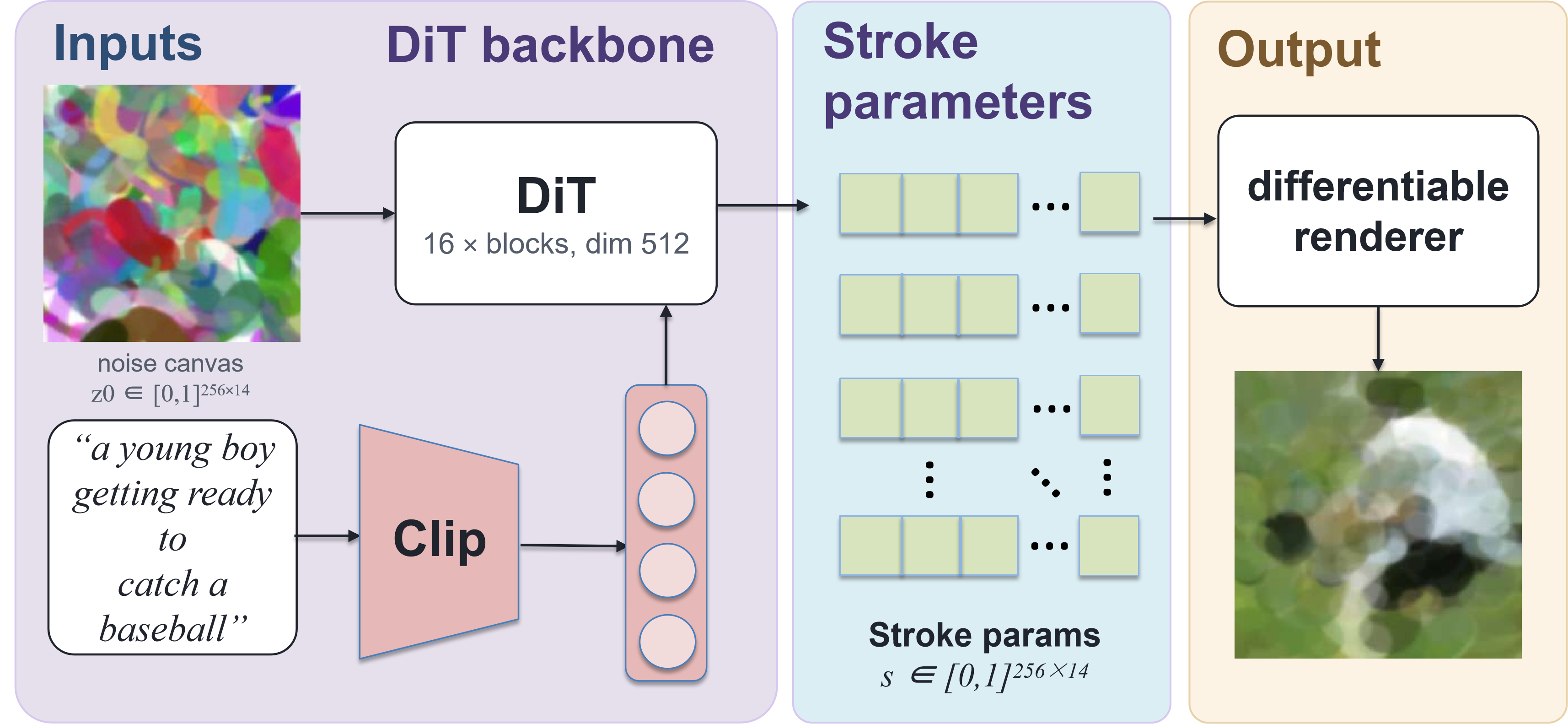}
\caption{Overview. A noise canvas $z_0\in[0,1]^{256\times14}$ and a frozen CLIP
text embedding condition a DiT that outputs $256$ stroke parameter vectors of width
$14$; a differentiable renderer composites the $256$ B\'{e}zier curves into the
image in a single forward pass. Training supervises the rendered image with a
pixel-space MSE loss whose gradient flows back through the renderer to the stroke
parameters and the DiT (Sec.~\ref{sec:generator}).}
\label{fig:pipeline}
\end{figure*}

Our generator describes an image the way a draftsman does, as a small set of strokes, and renders those strokes to pixels
(Fig.~\ref{fig:pipeline}).

\subsection{Stroke representation and differentiable renderer}
\label{sec:strokes}

\begin{figure}[tb]
\centering
\includegraphics[width=0.6\columnwidth]{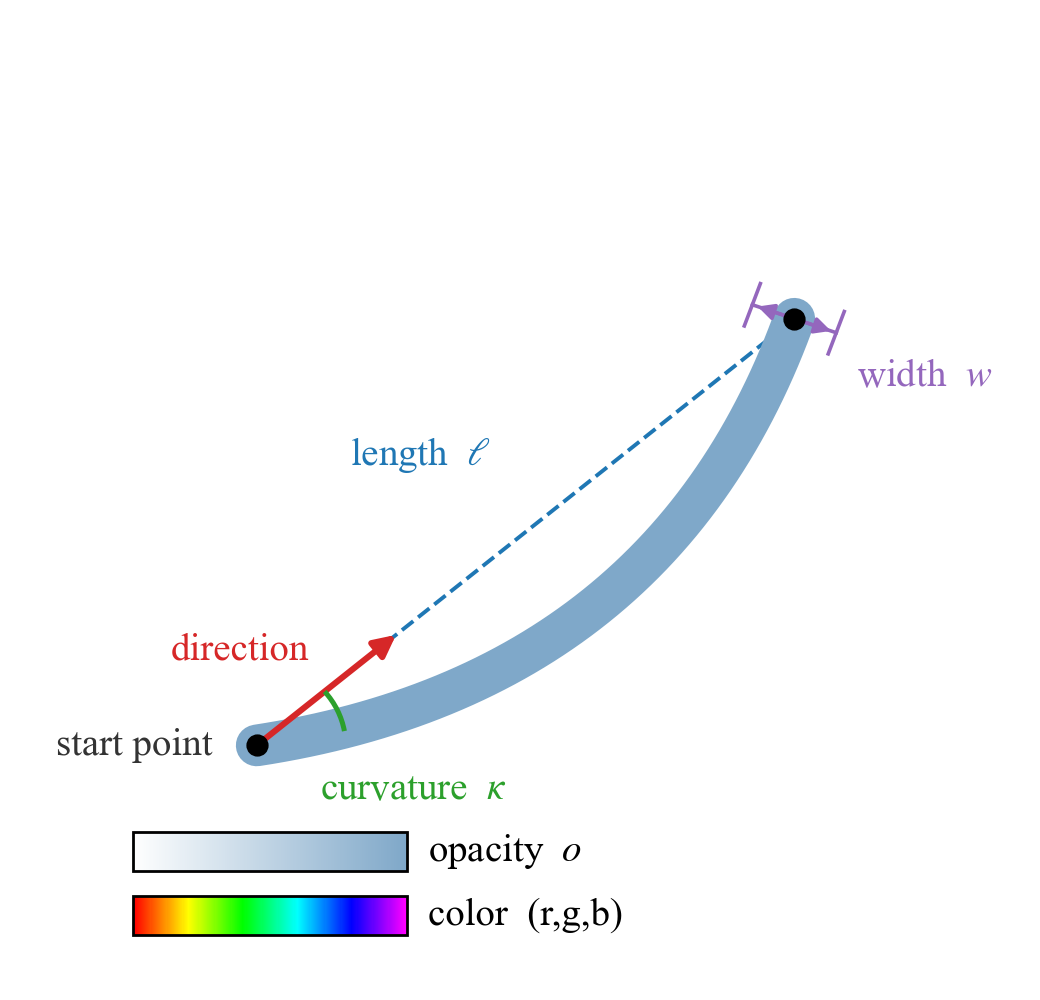}
\caption{A single sdlc stroke, illustrating its five geometric parameters:
start point, direction, length, width, and curvature (the dashed line is the
straight chord, so the gap to the curve is the curvature). Color and opacity set
its appearance. The full parameter layout is in Table~\ref{tab:stroke_params}.}
\label{fig:stroke_geom}
\end{figure}

A scene is described by $256$ cubic B\'{e}zier strokes. A pixel records a colour at
a location; a stroke records a mark, with a start, a direction, a weight and a
colour, so the output is a list of drawing decisions rather than a grid of
values. Each stroke is parameterized by $11$ numbers (Table~\ref{tab:stroke_params}): a start point, a
unit direction, a length, a curvature, a width, an RGB color, and an opacity
(Figure~\ref{fig:stroke_geom}). Given the start point and the direction times
length, the two endpoints are fixed; the curvature bows the control handles
laterally; together these determine the four B\'{e}zier control points. The
input noise tensor has width $14$: slots $0$--$10$ map to the eleven renderer
parameters, and slots $11$--$13$ are free latent channels injected into the
token representation but not read by the renderer. These three channels are
load-bearing, not idle padding: the eleven semantic parameters are coupled, leaving
few degrees of freedom for per-stroke information not already fixed by the
geometry, and zeroing the free channels at inference collapses training-set PSNR
from $19.27$ to $7.24$\,dB (Sec.~A).

\begin{table}[tb]
\centering
\small
\caption{The 11 semantic parameters of a stroke and their roles in the
renderer. The start point plus the direction-scaled length fix the endpoint;
the curvature bows the control handles. Slots $11$--$13$ are free latent
channels: the renderer ignores their output, but they are mixed by the input
projection and the model relies on them (Sec.~\ref{sec:strokes}).}
\label{tab:stroke_params}
\vspace{2pt}
\fitwidth{%
\begin{tabular}{cll}
\toprule
Slot & Parameter & Controls \\
\midrule
0--1  & start point $(p_x,p_y)$ & where the stroke begins \\
2--3  & direction $(d_x,d_y)$ & unit vector of approach \\
4     & length $\ell$ & stroke extent \\
5     & curvature $\kappa$ & lateral bow (straight..arc) \\
6     & width $w$ & stroke thickness (px) \\
7--9  & color $(r,g,b)$ & stroke color \\
10    & opacity $o$ & alpha-blending weight \\
\midrule
11--13 & free latent channels & extra degrees of freedom \\
\bottomrule
\end{tabular}}

\end{table}

The renderer evaluates, for every pixel, its signed distance to each stroke
centerline, converts that distance into a soft coverage with a sigmoid whose
sharpness tracks the half-width, and composites the strokes sequentially with
alpha blending, stroke $k$ painted over the strokes before it.

Gradients reach the strokes through four factors. Write $L$ for the pixel MSE to
the target $I^*$, $\theta_k$ for the eleven semantic parameters of stroke $k$,
$I^{(k)}$ for the canvas after the first $k$ strokes, and $K=256$ for the number
of strokes; $\alpha_{k,ij}$, $d_k$ and $\mathbf{P}_j$ are stroke $k$'s coverage,
its distance, and its $j$-th control point as before. Then
\begin{equation}
  \frac{\partial L}{\partial \theta_k}
  = \sum_{i,j} \frac{\partial L}{\partial \alpha_{k,ij}}\,
      \frac{\partial \alpha_k}{\partial d_k}\,
      \frac{\partial d_k}{\partial \mathbf{P}_j}\,
      \frac{\partial \mathbf{P}_j}{\partial \theta_k},
\end{equation}
in which coverage carries the occlusion of every stroke painted after $k$,
\begin{equation}
  \frac{\partial L}{\partial \alpha_{k,ij}}
  = \frac{2}{3HW}\big(I^{(K)}_{ij}-I^{*}_{ij}\big)^{\!\top}
    \big(\mathbf{c}_k-I^{(k-1)}_{ij}\big)
    \prod_{r=k+1}^{K}\big(1-\alpha_{r,ij}\big),
\end{equation}
coverage is a sigmoid of the signed distance,
$\partial\alpha_k/\partial d_k=-(o_k/\tau)\,\sigma(u_k)(1-\sigma(u_k))$ with
$u_k=(w_k/2-d_k)/\tau$, the distance moves along the unit normal at the closest
point as
$\partial d_k/\partial\mathbf{P}_j=\binom{3}{j}(1-t_k)^{3-j}t_k^{\,j}\mathbf{n}_k$,
and the map from the eleven semantic parameters to the four control points is
closed form, so $\partial\mathbf{P}_j/\partial\theta_k$ is a constant Jacobian
(Sec.~B derives each factor). This is what makes the structural prior
trainable: the loss is still pixel MSE, but it now reaches eleven interpretable
parameters per stroke.

\subsection{The generator}
\label{sec:generator}

The generator is a DiT~\cite{peebles2023scalable} with $16$ blocks of dimension
$512$ and rotary position embeddings~\cite{su2021roformer}, $67.83$M parameters in total. Its input is a noise
canvas $z_0\in\mathbb{R}^{256\times14}$, drawn from
$\mathrm{Uniform}[0,1]^{256\times14}$ for every training sample; each of the $256$
rows is one stroke slot. Rendering a raw $z_0$ yields scribbles; because the canvas is re-drawn per sample,
the stroke parameters must follow the text. A learned per-stroke
embedding distinguishes the slots, which matters because the tokens are otherwise
exchangeable and attention would average them. A frozen CLIP encoder~\cite{rombach2022high} ($77\times768$) supplies the
condition through cross-attention in every block. The output head projects each of the $256$
tokens to stroke parameters, which the renderer composites into a
$256\times256$ image. The training loss is the pixel MSE between the rendered
image and the ground truth,
\begin{equation}
  L_{\mathrm{pictura}} = \frac{1}{3HW}\sum_{i,j,c}\left(R(z_0,c)_{i,j,c} - I^*_{i,j,c}\right)^2,
\end{equation}
where $R(z_0,c)$ is the image rendered from $z_0$ and the text condition $c$,
$I^*$ the target image, and $H,W$ and $i,j,c$ the spatial and channel indices.

The model is a single conditional map from a noise canvas and the text to stroke
parameters, with no sampling loop: one forward pass produces the whole image. It composes the drawing in a single
pass rather than assembling it over many. Like
the diffusion baselines below, it starts from noise and must produce an image.

Different canvas draws select different stroke sets and change the picture: on
eight draws of one caption every semantic slot moves and $88$--$94\%$ of pixels
change by more than $0.05$, while each draw still follows the caption
(Sec.~C). The free channels are the exception (they stay fixed), so the text
fixes \emph{what} is drawn and the canvas selects \emph{which} valid configuration
draws it, giving image-level variation, not stroke-level jitter.

\section{Experiments}
\label{sec:experiments}

\paragraph{Dataset.}
We use three corpora, each of which answers a different question. All are rendered
or cropped to $256\times256$.
\begin{itemize}
  \item \textbf{COCO}~\cite{lin2014microsoft}: a $1{,}000$-image sample of real
  photographs, each with its first caption. Since one caption names one image, this
  corpus measures whether the model reproduces the image a caption describes, scored
  by PSNR; all main-text numbers come from it, on a fixed $50$-prompt subset.
  \item \textbf{multi100}: $100$ templates $\times$ $10$ Stable-Diffusion renders
  each. The first $80$ templates are used for training and the last $20$ are held
  out; KID~\cite{binkowski2018demystifying} is computed on generated samples against
the real images of the same
  split. This corpus measures whether the model covers the concept, not whether it recalls one image.
  \item \textbf{SD gen}: $1{,}000$ images rendered by Stable Diffusion, one per
  templated caption, so the caption is the only source of variation; both arms are
  trained and scored on it, which makes this a within-corpus comparison away from COCO, not a zero-shot test (Sec.~D, Sec.~\ref{sec:gen}).
\end{itemize}
All images are center-cropped to a square, resized to $256\times256$ with bilinear
interpolation, and normalized to $[0,1]$. Both the stroke and the pixel arms train
on the same images and are scored with the same metric on the same prompt set.

\paragraph{Implementation details.}
Adam at learning rate $10^{-4}$ and weight decay $10^{-4}$ for the stroke model,
AdamW~\cite{loshchilov2019decoupled} for the pixel baseline; both use a half-cycle cosine schedule to
$0.05\times$ peak. The stroke model uses batch size $4$ with $4$ accumulation steps
and clipping at $1.0$. The generators train for $2{,}000$ epochs on the
$1{,}000$-image set, the regressions for $450$; every comparison below is on
training prompts.

\paragraph{Baselines.}
We compare against pixel-output models of comparable size, all predicting the full
$196{,}608$-value image. \textbf{Pixel diffusion} is a standard U-Net denoiser~\cite{ronneberger2015unet} with
$74.63$M parameters, trained with a DDPM schedule~\cite{ho2020denoising} and the de-facto standard
text-to-image recipe (v-prediction, min-SNR-$\gamma{=}5$
weighting~\cite{hang2023efficient}, scaled-linear noise, EMA $0.999$,
classifier-free guidance at $w{=}3$~\cite{ho2022classifier}, $50$ DDIM
steps~\cite{song2020denoising}); its channel
widths are $96/192/384/384$ with two layers per block and cross-attention to the
same frozen CLIP features. A $35\times35$ variant uses the identical recipe and
U-Net, changing only the output resolution ($3{,}675$ values) to match Pictura's
dimension. \textbf{Pixel regression} is the direct pixel counterpart of Pictura: the
same DiT trunk, text conditioning, noise canvas and image-space MSE loss, with the
stroke head replaced by a linear head that emits a raster; we train it at
$256\times256$ and $36\times36$ (output dimension $3{,}888\approx$ Pictura's
$3{,}584$). Each costs one forward pass at inference, like Pictura, and each draws
its own noise sample per image.

\paragraph{Evaluation protocol and why PSNR.}
All final comparisons (Table~\ref{tab:main}) score every model on the same $50$
training prompts against $256\times256$ ground truth with one sampler and seed, and
report mean $\pm$ std over five training runs; low-resolution outputs are shown
upsampled. We judge by PSNR because it is the task
objective, not the diffusion training objective. Because PSNR alone conflates
binding with fidelity, and a one-step model can raise it by memorizing, we always
read it alongside a two-input sensitivity probe (Sec.~\ref{sec:binding}) and the
qualitative panels (Fig.~12).

\paragraph{Setup and fairness.}
Both arms share the COCO $1{,}000$ images, the same $50$ scored captions, and the
same frozen CLIP ViT-L/14 (SD-1.5) encoder; they differ in parameters ($67.83$M
versus $74.63$M, the nearest configuration the U-Net admits, with the stroke model
the smaller), output dimensions, and loss. The per-image baseline in every table is
the constant-mean image computed on the prompt pool that table is scored on.

\subsection{Main result}
\label{sec:main}

\begin{figure*}[tb]
\centering
\includegraphics[width=\textwidth]{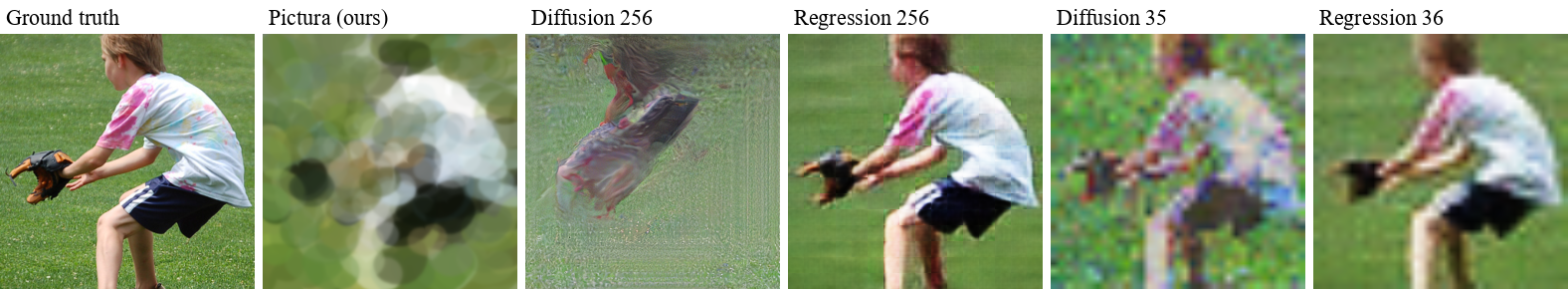}
\caption{A representative training example. From left to right: ground truth,
Pictura, standard-recipe diffusion ($256\times256$), pixel regression
($256\times256$), the dimension-matched diffusion ($35\times35$, output dim
$3{,}675\approx$ Pictura's $3{,}584$; upsampled to $256$), and pixel regression
at $36\times36$ (output dim $3{,}888$, upsampled). The two regressions share
Pictura's DiT trunk, text feature and noise canvas and differ only in the output
head; all six models see the same prompt. Pictura produces a full-resolution image
that visibly follows the prompt; the $256\times256$ diffusion renders a noisy,
prompt-specific scene far from the target pixel-wise; the $35\times35$ column
varies with the prompt only at very low resolution. The regressions reach higher
PSNR at the same one-step cost, which is the point Sec.~\ref{sec:binding}
interrogates.}
\label{fig:teaser}
\end{figure*}

Table~\ref{tab:main} compares Pictura against a standard-recipe text-to-image diffusion model ($50$ DDIM steps) on the same $50$ prompts. The ordering it reports is not the usual one: at full resolution the diffusion scores \emph{below} the constant-mean image ($12.01$ against $12.51$\,dB), while the stroke model clears it by $6.62$\,dB.
Figure~\ref{fig:teaser} shows one representative case; the marks it places stay visible as strokes, so the image reads as a drawing of the scene rather than as a pixel field that approximates it. The two regressions are close to the target in PSNR; what they lose is the
response to their noise, which Sec.~\ref{sec:binding} isolates.

\begin{table}[tb]
\centering
\small
\caption{PSNR against $256\times256$ ground truth, mean $\pm$ std over five
training runs (two extra seeds per arm, plus the reference run); the std measures
sensitivity to initialization, not to prompts. The two regression rows are one-step
positive controls, excluded from the pairwise bootstrap because their score measures recall, not generation (Sec.~\ref{sec:binding}). They share Pictura's DiT
trunk and differ only in the output head ($67.88$M at $36\times36$, $68.22$M at
$256\times256$), trained for $450$ epochs. Sec.~P repeats the
comparison on images no model has seen.}
\label{tab:main}
\vspace{2pt}
\fitwidth{%
\begin{tabular}{lccc}
\toprule
Method & Par. & Out.dim & PSNR \\
\midrule
\textbf{Pictura (stroke DiT)}                    & $67.83$M & $3{,}584$   & $\mathbf{19.13 \pm 0.12}$ \\
Pixel diffusion, standard recipe ($50$ steps)    & $74.63$M & $196{,}608$ & $12.01 \pm 0.09$ \\
\midrule
Pixel diffusion, $35\times35$ ($50$ steps)       & $74.63$M & $3{,}675$   & $16.40 \pm 0.04$ \\
Pixel regression, $36\times36$ (1 step)          & $67.88$M & $3{,}888$   & $21.88 \pm 0.01$ \\
Pixel regression, $256\times256$ (1 step)        & $68.22$M & $196{,}608$ & $25.20 \pm 0.29$ \\
\midrule
Constant-mean baseline                           & ---      & ---         & $12.51$ \\
\bottomrule
\end{tabular}}

\end{table}

One entry does not fit: the $256\times256$ regression scores higher than any generator. We return to it in Sec.~\ref{sec:binding}, because until it is explained no row of this table can be read as a generation score.

Because every model is scored on the same prompts, the comparison is paired.
Table~\ref{tab:paired} resamples the prompts with replacement: against the
dimension-matched $35\times35$ diffusion Pictura leads by $2.93$\,dB ($95\%$
interval $[2.02, 3.91]$) and wins on $45$ of $50$ prompts; against the
full-resolution diffusion it leads by $7.34$\,dB and wins on every prompt, so the
gap is not an artefact of a few outliers. Per-prompt values are in Fig.~14.

\begin{table}[tb]
\centering
\small
\caption{Paired comparison on the shared $50$ prompts, computed on the reference
run (the first of the five training runs behind Table~\ref{tab:main}), since a
paired test needs one prediction per prompt per model. $\Delta$ is the mean
per-prompt difference (Pictura minus baseline); the interval is a
$10{,}000$-sample bootstrap over prompts. ``Wins'' counts prompts on which
Pictura scores higher.}
\label{tab:paired}
\vspace{2pt}
\fitwidth{%
\begin{tabular}{lccc}
\toprule
Comparison & $\Delta$ PSNR & $95\%$ CI & Wins \\
\midrule
vs.\ diffusion $35\times35$ & $+2.93$ & $[+2.02, +3.91]$ & $45/50$ \\
vs.\ diffusion $256\times256$ & $+7.34$ & $[+6.68, +7.99]$ & $50/50$ \\
\bottomrule
\end{tabular}}

\end{table}

At inference Pictura is cheaper at every stage (Table~\ref{tab:time}): $6.3$\,ms per
forward pass against $72.2$\,ms for a single DDIM step at $256\times256$, and
$37.1$\,ms end-to-end against $3604$\,ms for the $50$-step model. Its $30$\,ms of
stroke compositing is small compared with the sampling it replaces (Sec.~E).

\begin{table}[tb]
\centering
\small
\caption{Per-image inference time (ms) on an NVIDIA RTX PRO 4000 Blackwell,
randomly initialised weights, mean over $10$ images (std within $0.5$\,ms).
``Forward'' is the network only; ``end-to-end'' includes rendering/compositing to
$256\times256$. Pictura and the DiT pixel regressions share the trunk and differ
only in the output head, so their forward cost is the same; what Pictura adds is
$30$\,ms of stroke rendering. Fifty DDIM steps cost $3604$\,ms at full resolution
and $838$\,ms at $35\times35$, so the step count, not the trunk, sets the diffusion
cost.}
\label{tab:time}
\vspace{2pt}
\fitwidth{%
\begin{tabular}{lcc}
\toprule
Method & forward (ms) & end-to-end (ms) \\
\midrule
\textbf{Pictura (1 step)}                     & $\mathbf{6.3}$  & $\mathbf{37.1}$ \\
Pixel regression $256$ (DiT, 1 step)          & $6.3$ & $6.3$ \\
Pixel regression $36$ (DiT, 1 step)           & $6.3$ & $6.4$ \\
\midrule
Diffusion $256$ (50 steps)                    & $3604.4$ & $3608.5$ \\
Diffusion $256$ (1 step)                      & $72.2$ & $72.4$ \\
Diffusion $35$ (50 steps)                     & $837.9$ & $839.6$ \\
Diffusion $35$ (1 step)                       & $17.2$ & $17.0$ \\
\bottomrule
\end{tabular}}

\end{table}

\subsection{A drifting-field comparison}
\label{sec:drift}

Every comparison so far pits a deterministic decoder against a denoising
objective. An arm trained on per-pixel MSE is pulled toward the conditional mean of
the targets, so it is a blur by construction and any distribution distance will
penalize it whatever its conditioning; the diffusion arm samples; it does not average. A KID comparison between the two therefore confounds the output space with
the training objective. To separate them we also train with a drifting
field~\cite{deng2025drifting}, which matches distributions directly, is one-step like
ours, and needs no discriminator. For a
generated sample $x$ it pulls
toward the batch's targets $p_i$ and pushes from the other samples $x_j$,
\begin{equation}
\begin{aligned}
  V(x) &= \sum_i w^{+}_i(x)\,p_i-\sum_j w^{-}_j(x)\,x_j,\\
  w^{\pm}_i &\propto e^{-\|x-\cdot\|^2/2\tau^2},\qquad
  L_{\mathrm{drift}}=\mathbb{E}_x\|V(x)\|^2,
\end{aligned}
\end{equation}
summed over three kernel widths and normalized per sample (Sec.~F). Because
the field is a distance it can be computed in any space, and the choice of space
turns out to matter more than the field itself. The three drifting arms share the
multi100 corpus and the $68$M DiT backbone; the diffusion row is the
parameter-matched $74.63$M U-Net (Table~\ref{tab:drift}).

\begin{table}[tb]
\centering
\small
\caption{KID on the held-out captions at the final epoch, so the three
drifting arms are compared under the same checkpoint criterion. Intervals are $95\%$
bootstraps resampling the $20$ held-out captions, with $10$ generated samples per
caption against its $10$ targets; the stroke and pixel intervals are adjacent
($0.292$ against $0.293$), though the ordering is stable across four generation
seeds. The other three arms are not scored here, for the reason given in
Sec.~F.}
\label{tab:drift}
\vspace{2pt}
\fitwidth{%
\begin{tabular}{lcc}
\toprule
Field space / method & final KID & 95\% CI \\
\midrule
Stroke parameters        & $\mathbf{0.264}$ & $[0.251, 0.292]$ \\
Raw pixels               & $0.308$ & $[0.293, 0.334]$ \\
ResNet features          & $0.420$ & $[0.394, 0.451]$ \\
\bottomrule
\end{tabular}}

\end{table}

Stroke-parameter space reaches the lowest final KID ($0.264$, CI $[0.251,0.292]$),
ahead of raw pixels ($0.308$) and the ResNet features used in
\citeauthor{deng2025drifting} ($0.420$); the intervals are adjacent, but the
ordering holds at four generation seeds. More telling is the trajectory: every arm
peaks early and degrades while its loss keeps falling, most sharply for the diffusion arm, because once a sample-level objective can be driven toward zero on
$1{,}000$ images, the fastest way down is to reproduce targets instead of the distribution (Sec.~F).

\subsection{Where the gap comes from}
\label{sec:dimension}
\label{sec:undertrained}

\begin{figure}[tb]
\centering
\includegraphics[width=\columnwidth]{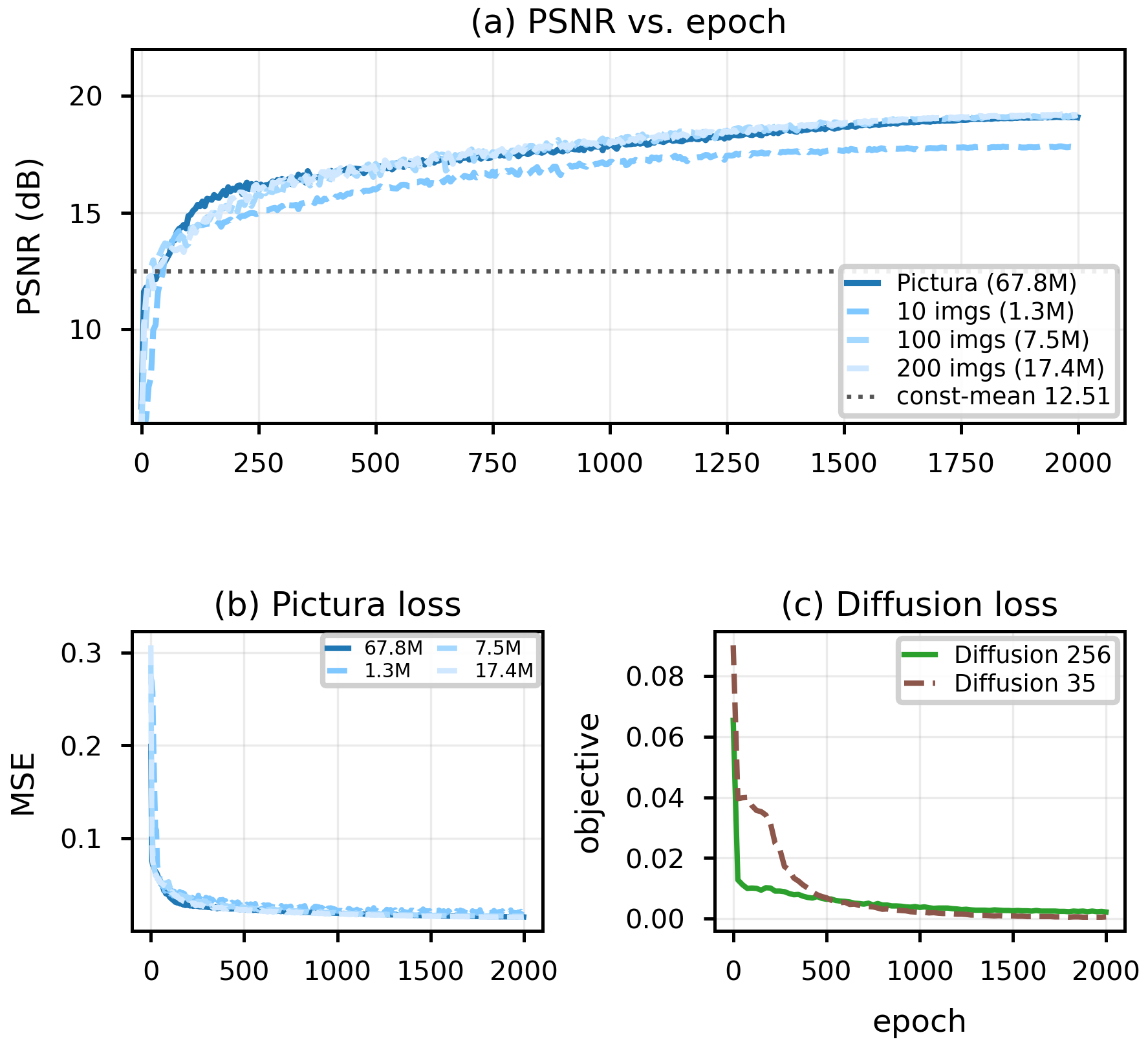}
\caption{(a) PSNR against epoch for Pictura and for three smaller-scale variants, the lighter curves being the $10$, $100$ and $200$-image runs in that order: every run rises past the constant-mean baseline ($12.51$\,dB, dotted) and continues to the high teens. We do not plot a PSNR curve for the diffusion
baselines because PSNR is not their training objective. (b) Pictura's
image-space MSE loss falls throughout. (c) The pixel baselines' own denoising
objectives also fall, yet their PSNR never rises, showing they fit a solution that
does not track the prompt.}
\label{fig:curves}
\end{figure}

\paragraph{Training length.}
The standard recipe was trained to the full $2{,}000$ epochs, and its \emph{training-set} PSNR plateaus at $12.01$\,dB while its training loss falls from $0.0056$ to $0.0022$, the score moving $0.1$\,dB over the last $1{,}350$: the signature of a model fitting its objective, not the text. Figure~\ref{fig:curves} shows the
contrast: Pictura's PSNR crosses the baseline within tens of epochs and continues to
$19.13$\,dB (the three scaling runs follow the same schedule), while both models
keep decreasing their own losses and only Pictura's score rises.

\paragraph{Sampling budget.}
On the same $50$-prompt subset, reducing the sampler
from $50$ DDIM steps to one drops PSNR from $12.01$ to $6.85$\,dB at $256\times256$
and from $16.40$ to $8.76$\,dB at $35\times35$, so truncation cannot explain the gap
either. A truncated sample can look recognizable yet score low, since PSNR penalizes
the residuals from which structure is reconstructed.

\paragraph{The output.}
What remains is the output itself, which differs in two ways: dimensionality and structure. Lowering the output resolution by a factor of seven \emph{improves} the score by $4.4$\,dB: shrinking the pixel output to $35\times35$ ($3{,}675$ dimensions, close to Pictura's $3{,}584$) raises its PSNR from $12.01$ to $16.40$, clearing the baseline. A low-dimensional pixel output therefore recovers much of the condition.

Dimension alone does not close the gap. At the same $\sim3.5$k dimensions the
structured stroke output reaches $19.13$, $2.7$\,dB above the low-dimensional
pixel baseline. Both sit in the same low-dimensional regime, so the remaining gap
is attributable to structure: a stroke parameter has a bounded role and the set of
valid strokes is a thin manifold, whereas a $35\times35$ pixel output has neither.
The stroke representation therefore addresses both factors at once, in a single
forward pass, whereas the low-dimensional pixel diffusion needs $50$ denoising
steps to reach $16.40$ (see the sampling budget above).

The reachable cells are listed in Sec.~I: three corners of a
$2\times2$ design plus a regression control. Its fourth cell (a $36\times36$
raster emitted directly by the same DiT backbone, data, and per-sample noise draw,
with no renderer and no denoising loop) is what controls for structure at low
dimension. It reaches $21.88$\,dB, above both generators, so dimension and training
signal alone do raise the score; but the probe of Sec.~\ref{sec:binding} shows that
score is recall (noise diversity $0.002$, prompt-to-noise ratio $134$), so it does
not recover the condition. The comparison that matters for generation is between
models that respond to their noise, and there the stroke arm leads at matched
dimension in a single step.

The design is not a strict $2\times2$: a stroke costs $14$ values and a pixel $3$, so the two cannot both reach $196{,}608$, and $256$ strokes already saturates (Sec.~G). At $256\times256$ the pixel diffusion does not
produce a coherent picture: its samples are fragmented scenes that fall below the
constant-mean baseline. A coherent picture appears only at $35\times35$, and that
coherence is bought with resolution, since $3{,}675$ values cannot carry the detail
of a $256\times256\times3$ image and upsampling does not restore it. Pictura is
coherent at full resolution: $3{,}584$ parameters render a $256\times256$ image.

\subsection{Conditional binding and fidelity}
\label{sec:binding}

PSNR alone does not separate whether a model binds the condition from
how faithfully it renders. We measure both: prompt sensitivity (we fix the
noise seed, change only the prompt, and report the mean absolute difference
between outputs, averaged over prompt pairs) and the PSNR, for every model under
the same protocol.

\begin{table}[tb]
\centering
\small
\caption{Prompt and noise sensitivity. The ``Space'' column names what the
difference is measured on, and only rows with the same entry are directly
comparable: an L1 distance between $256\times14$ stroke parameters is not the same
unit as one between $256\times256$ images, which is why Pictura appears twice.
Prompt diff.\ is the mean pairwise absolute output change across $8$ captions at a
fixed noise draw; noise diff.\ the mean pairwise change across $8$ noise draws at a
fixed caption. On the rendered image, the basis every row except the stroke-space
one uses, Pictura responds to the prompt most ($1.92$ against $1.31$ and
$1.04$); in its own parameter space its ratio is higher still ($3.07$), reported separately, not mixed into the image-space ranking.}
\label{tab:binding}
\vspace{2pt}
\fitwidth{%
\begin{tabular}{lcccc}
\toprule
 & Prompt diff. & Noise diff. & Ratio & Space \\
\midrule
\textbf{Pictura (1 step)}            & $\mathbf{0.267}$ & $0.139$ & $\mathbf{1.92}$ & image \\
\textbf{Pictura}                     & $0.249$ & $0.081$ & $3.07$ & stroke \\
Pixel diffusion, $35\times35$        & $0.136$ & $0.109$ & $1.25$ & image \\
Pixel diffusion, $256\times256$      & $0.124$ & $0.122$ & $1.02$ & image \\
Pixel regression, $36\times36$       & $0.293$ & $0.002$ & $134$  & image \\
Pixel regression, $256\times256$     & $0.335$ & $0.002$ & $164$  & image \\
\bottomrule
\end{tabular}}

\end{table}

The picture is graded rather than binary. On the rendered image, where all rows are
comparable, Pictura binds most strongly ($0.267$, ratio $1.92$), the $35\times35$
pixel diffusion is weaker ($0.136$, $1.25$), and the $256\times256$ diffusion is
weaker still ($0.124$, $1.02$), set by its noise, not its text. Pictura's
stroke-space ratio ($3.07$) is reported separately, since a parameter distance and a
pixel distance are not the same quantity. The regressions' higher prompt difference
comes only from their output being a function of the caption alone; their noise
difference of $0.002$ shows this is not binding.

\paragraph{Two failure modes, one table.}
Table~\ref{tab:binding} separates the two ways a one-step model can go wrong here. Pixel diffusion at $256\times256$ is a working sampler set by its noise (ratio $1.02$). Regression tops the PSNR table by emitting a deterministic function of the caption (noise $0.002$, ratio $164$), the simplest way to lower pixel MSE on a one-to-one corpus. We therefore read PSNR as a generation score only where the noise column is non-negligible and the ratio is well above one.
How far any of this generalizes is a separate question. On $60$ images that no model
has seen, every arm falls to $9.5$--$9.8$\,dB, a $0.35$\,dB spread where the
training set showed $13$\,dB, and the two arms that lead on the training corpus fall
furthest (Sec.~P). With $1{,}000$ images reconstruction does not carry over,
so these columns measure binding on the training prompts, which is what the probe is
for.

\subsection{Generation versus reconstruction}
\label{sec:genrec}

\begin{figure}[tb]
\centering
\includegraphics[width=\columnwidth]{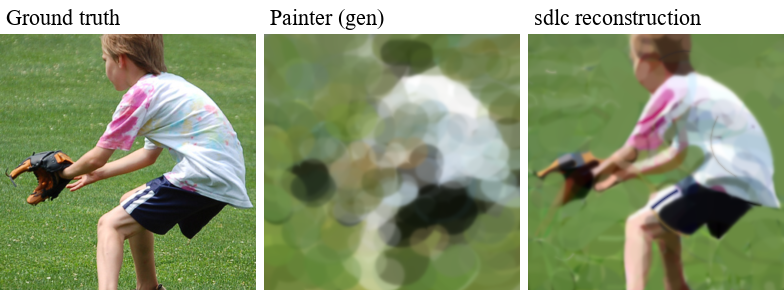}
\caption{Generation versus reconstruction on the same image. Left: ground truth.
Middle: Pictura ($19.13$\,dB). Right: per-image gradient optimization of the same
$256$ sdlc strokes with no network ($24.40$\,dB). The representation can
reconstruct far better than one amortized pass achieves; the gap measures the cost
of one-step generation, not the limit of the output.}
\label{fig:genrec}
\end{figure}

The representation has headroom that the generator does not use. Per-image optimization of the same $256\times14$ parameters, through the same
renderer and with no network, reconstructs the baseball image at $24.40$\,dB
(Fig.~\ref{fig:genrec}), so the representation carries more than one amortized pass
reaches. The step count does not explain the gap either: the fifty-step diffusion
reaches only $12.01$\,dB, below the constant-mean baseline. Both arms are limited by the $1{,}000$-image budget rather than by their objective, since more epochs do not help (Sec.~\ref{sec:undertrained}), and the $5.3$\,dB between $19.13$ and $24.40$ is capacity a single pass has yet to claim.

Because the output is a set of curves rather than a raster, post-hoc optimization
on the stroke parameters raises fidelity without regenerating the image
(Sec.~H). We claim parameter-level, not object-level, interpretability: which
object a stroke depicts is neither annotated nor supervised.

\subsection{A generalization probe on SD gen}
\label{sec:gen}

The corpora above all reward reproducing a single image per caption. On SD gen, a strict one-to-one corpus of $1{,}000$ pairs whose captions are templated, we train both arms on the same pairs and score them on the same set, a within-corpus comparison on a distribution away from COCO, not a zero-shot test. Pictura reaches $18.8$\,dB against a constant-mean
baseline of $13.1$\,dB, below its COCO score; the $256\times256$ pixel regression,
trained on exactly the same pairs, reaches $24.18$\,dB.

The pixel arm wins here because templated captions make the map more regular than COCO's, so both arms fit it easily: a simpler task is easier to memorize, which is not a statement about the representation. With one image per caption, neither number is open-vocabulary generalization (Sec.~D).

\section{Discussion}
\label{sec:discussion}

The field treats the output space as given. Text-to-image models inherit the
raster and spend their capacity on the generator that fills it. We keep the
generator and replace the raster: $256$ B\'{e}zier strokes, $3{,}584$ numbers
instead of $196{,}608$. Holding data, parameters, text encoder and training length
fixed, we move only that inherited choice, and at $1{,}000$ images it changes the
result: $19.13$\,dB against $12.01$\,dB for a comparable diffusion
model, and $2.7$\,dB at matched dimension, in one pass rather than fifty.

A stroke is not a coordinate with a value: it has a bounded role. The renderer maps its eleven parameters onto a thin manifold of valid strokes, so the network never has to represent the set of plausible pixels, only to decide which strokes to place. The loss is still pixel MSE and the gradient still flows through
the image; what changes is where it lands, on eleven interpretable parameters per stroke, not on $196{,}608$ coordinates. Because the prior lives in the
decoder, every forward pass re-applies it for free.

Two consequences follow from this. One pass carries the whole image: $6.3$\,ms per forward
against $72.2$\,ms for a single DDIM step, $37.1$\,ms against $3{,}604$\,ms for fifty steps (Sec.~E).
The parameters also stay legible to a distribution metric that never sees a stroke:
the drifting field reaches its lowest KID in stroke-parameter space ($0.264$), ahead of
pixel space ($0.308$) and ResNet features ($0.420$).

The instructive counterexample is the one-step pixel regression. It scores higher
on PSNR and learns a deterministic function of the caption, its output moving by
$0.002$ when the noise changes and by $0.335$ when the caption does. The difference is like that between a sketched copy and an impressionist painting: the copy reproduces every detail of its subject, while the painting composes the picture from an impression. The regression reproduces the image it has seen, whatever noise it starts from; Pictura puts down strokes, and the same caption gives a different picture on each canvas. At this data scale PSNR
separates the arms only after the probe has told us what they are doing; the probe,
not the score, is what turns a number into evidence about conditioning.

\subsection{Limitations}
\label{sec:limits}

We report a single resolution and a small set of corpora; whether the advantage
holds at higher resolution or with more strokes is untested. Every arm is trained
five times and the spread is reported with each mean in Table~\ref{tab:main}; the
intervals in Tables~\ref{tab:paired} and~\ref{tab:drift} are paired bootstraps. On unseen images all five
arms land within $0.35$\,dB of each other (Sec.~P), so our claims
concern conditioning on the training corpus, not held-out synthesis. The generator trains from scratch, and fine-tuning a pretrained text encoder would change the data-efficiency picture. Our primary measure is conditioning recall, not perceptual quality: Sec.~J adds LPIPS, CLIPScore and FID, of which FID follows the
PSNR ordering, LPIPS supports only the $256\times256$ comparison, and CLIPScore
does not separate the models. 

\section{Conclusion}
\label{sec:conclusion}

We presented Pictura, a one-step text-to-image generator whose output is a set of
$256$ B\'{e}zier strokes rather than a pixel grid, together with a controlled study
of what that choice buys. Under a shared data, parameter, text-encoder and training
budget, the stroke decoder reaches $19.13$\,dB on $1{,}000$ COCO images where pixel diffusion reaches $12.01$, keeps a $2.7$\,dB lead at matched
output dimension, and produces an image $97\times$ faster than the fifty-step
diffusion it replaces.

The difference lies in the shape of the output space, not its size. A parameterized stroke restricts every sample to a thin manifold of valid strokes, and the renderer re-imposes that restriction on each forward pass, so the
model learns which strokes to place rather than how to resolve a raster. It is also what makes the output
editable stroke by stroke, which no pixel decoder offers at any PSNR.

Stroke output is therefore a direction worth developing, not a curiosity. It is cheap, editable by construction, and has absorbed almost none of the engineering that pixel models have.
More broadly, the output representation is a choice the field has not varied, and strokes are one setting of it.

In future work we hope to explore whether this effect survives at the scale of modern text-to-image systems, where no arm is memorizing, and whether giving each stroke object-level meaning can let stroke output convey objects and space as a human drawing does. In the end the work returns to an old intuition: a model should describe a scene with strokes, the way a person does, rather than arrange pixels in a grid.

\nocite{li2026omni,li2026unim}
\bibliographystyle{ieeenat_fullname}
\bibliography{main}

\clearpage
\appendix
\section{The three free latent channels}
\label{sec:suppl_freechan}

The noise width is $14$. Slots $0$--$10$ map to the eleven renderer parameters;
slots $11$--$13$ are free latent channels that are injected into the token
representation but are never read by the renderer. This section argues that they
are neither redundant nor optional.

\paragraph{Why the parameterization needs them.} The eleven semantic parameters are
coupled: start point, direction, length and curvature jointly determine the four
B\'{e}zier control points and overlap in what they control. That coupling leaves
few degrees of freedom for the network to encode per-stroke information that is not
already fixed by the geometry. The free channels supply an unconstrained
side-channel---information the model wants to route per stroke but that must not
corrupt the rendered image. They are free precisely because the renderer ignores
them, so the network can use them for internal computation without changing the
output.

\paragraph{Evidence that the model depends on them.} We run two checks: the mean
input-projection weight norm per slot group, and the training-set PSNR when the
free slots are zeroed at inference.

\begin{table}[h]
\centering
\small
\resizebox{\columnwidth}{!}{%
\begin{tabular}{lcc}
\toprule
Condition & mean $|W|$ on slots & training PSNR \\
\midrule
Unmodified                       & $0.283$ (slots $0$--$10$) & $19.27$ \\
Free slots $11$--$13$ zeroed     & $0.314$ (slots $11$--$13$) & $7.24$ \\
Semantic slots $0$--$10$ zeroed  & --- & $5.74$ \\
\bottomrule
\end{tabular}}
\caption{Free-channel ablation on a $50$-image subset of the training set. The
weight columns list the mean input-projection norm of each slot group in the
unmodified model; the PSNR column is measured with the named slots zeroed at
inference.}
\label{tab:freechan}
\end{table}

Two things follow from Table~\ref{tab:freechan}. First, the network allocates more
weight to the three free slots than to the eleven semantic ones
($0.314$ vs.\ $0.283$), so they are not unused capacity. Second, zeroing them
collapses the score to $7.24$\,dB---below the $12.51$\,dB constant-mean
baseline---which is the same order of failure as zeroing the semantic slots
outright ($5.74$\,dB). The channel is therefore load-bearing: the generator routes
work through it that the rendered image depends on.

\paragraph{Reading the subset.} The ablation runs on $50$ images for speed, so its
unmodified score ($19.27$) is measured on a slightly different subset of the
training set than Table~2 ($19.13$). The two should
not be compared directly; only the drop within this table is meaningful.

\paragraph{Relation to a fully semantic parameterization.} In a Bezier4-style
parameterization all $14$ dimensions would carry stroke meaning. That is a
different output head, not this one: here the renderer reads $11$ and the remaining
$3$ serve as a working channel the generator uses internally. Closing the gap
between the two is future work under a larger compute budget.

\section{The renderer in full}
\label{sec:suppl_renderer}

The main paper gives the four factors that carry the pixel loss back to a stroke's
semantic parameters; this section derives them in that order.

From the start point $\mathbf{p}_0$, the unit direction $\mathbf{d}$, and the
length $\ell$, the endpoint is $\mathbf{p}_1=\mathbf{p}_0+\ell\,\mathbf{d}$; the
two control handles are symmetric points offset laterally by the curvature
$\kappa$, yielding a cubic B\'{e}zier centerline
\begin{equation}
  B_k(t) = \sum_{j=0}^{3} \binom{3}{j}\,(1-t)^{3-j}\,t^{j}\,\mathbf{P}_j,
  \qquad t \in [0,1],
\end{equation}
where $\mathbf{P}_0,\mathbf{P}_1,\mathbf{P}_2,\mathbf{P}_3$ are the four control
points implied by the start point, direction, length, and curvature. The composite
image is built recursively over strokes, stroke $k$ painting on top of the canvas
accumulated from strokes $1$ through $k-1$,
\begin{equation}
  I^{(k)} = (1 - \alpha_k) I^{(k-1)} + \alpha_k \mathbf{c}_k,
  \qquad \alpha_k = o_k \, \sigma\!\left(\frac{w_k/2 - d_k}{\tau}\right),
\end{equation}
where $\mathbf{c}_k$ is the stroke color, $o_k$ its opacity, $w_k$ its width,
$d_k$ the pixel's distance to the centerline $B_k$, and $\tau$ the anti-aliasing
scale.

The first factor is the derivative of the loss with respect to the stroke's
coverage. Stroke $k$ is not confined to its own layer: every later stroke blends
over it, so differentiating the recursion gives
\begin{equation}
  \frac{\partial I^{(K)}_{ij}}{\partial \alpha_{k,ij}}
  = \big(\mathbf{c}_k - I^{(k-1)}_{ij}\big)
    \prod_{r=k+1}^{K}\big(1-\alpha_{r,ij}\big),
\end{equation}
an occlusion product over every stroke painted after $k$. With
$L=(1/3HW)\sum_{ij}\|I^{(K)}_{ij}-I^{*}_{ij}\|^{2}$ this becomes
$\partial L/\partial\alpha_{k}$ in the main paper. The second factor is the
coverage, a sigmoid of the signed distance,
\begin{equation}
  \frac{\partial \alpha_k}{\partial d_k}
  = -\frac{o_k}{\tau}\,\sigma(u_k)\big(1-\sigma(u_k)\big),
  \qquad u_k=\frac{w_k/2-d_k}{\tau}.
\end{equation}
The third is the signed distance's dependence on the control points: the closest
point $B_k(t_k)$ moves along the unit normal $\mathbf{n}_k$, and the centerline
moves along the Bernstein basis,
\begin{equation}
  \frac{\partial d_k}{\partial \mathbf{P}_j}
  = \binom{3}{j}(1-t_k)^{3-j}\,t_k^{\,j}\,\mathbf{n}_k .
\end{equation}
The fourth is the map from the eleven semantic parameters to the four control
points, which is closed form---the start point and the direction-scaled length fix
$\mathbf{P}_0$ and $\mathbf{P}_1$, and the curvature offsets $\mathbf{P}_2$ and
$\mathbf{P}_3$ laterally---so $\partial\mathbf{P}_j/\partial\theta_k$ is a
constant Jacobian. The four factors compose as
\begin{equation}
  \frac{\partial L}{\partial \theta_k}
  = \sum_{i,j} \frac{\partial L}{\partial \alpha_{k,ij}}\,
      \frac{\partial \alpha_k}{\partial d_k}\,
      \frac{\partial d_k}{\partial \mathbf{P}_j}\,
      \frac{\partial \mathbf{P}_j}{\partial \theta_k},
\end{equation}
with the sum over the pixels the stroke covers. Because the renderer is
differentiable end to end, this gradient is what lets the generator learn the
stroke parameters by backpropagation through the rendered image.

\section{Does the canvas actually change the output?}
\label{sec:suppl_diversity}

The generator takes a noise canvas as input, so one may ask whether a different
canvas produces a different output at all---plain pixel MSE against a single
target has the optimum $\mathbb{E}[I\mid c]$, which does not depend on the canvas,
and a model trained this way is free to ignore it. We therefore render one caption
eight times, resampling $z_0$ each time, and measure diversity at the level the
model actually outputs (the $256\times14$ stroke parameters) as well as on the
rendered image.

\begin{table}[h]
\centering
\footnotesize
\setlength{\tabcolsep}{4pt}
\begin{tabular}{lccc}
\toprule
Quantity (8 draws per caption) & ``monitor'' & ``surfer'' & ``sheep'' \\
\midrule
Stroke slots that move      & $256/256$ & $256/256$ & $256/256$ \\
Mean parameter difference   & $0.109$ & $0.091$ & $0.099$ \\
Image mean absolute diff.\  & $0.139$ & $0.101$ & $0.122$ \\
Pixels changing by $>0.05$  & $90\%$  & $88\%$  & $94\%$ \\
\bottomrule
\end{tabular}
\caption{Diversity under canvas resampling on three COCO captions---``A desktop
computer monitor sitting on top of a white desk'' (monitor), ``A surfer balances on
his board as he climbs the face of a wave'' (surfer), and ``A woman petting a sheep
on top of a green field'' (sheep). The caption is held fixed and $z_0$ is resampled
eight times; differences are means over the $28$ pairs of draws. The image itself
varies substantially---two thirds to nine tenths of its pixels change---not only
the stroke parameters.}
\label{tab:diversity}
\end{table}

The text fixes \emph{what} is drawn; the canvas selects \emph{which} stroke
configuration draws it. Crucially this is not stroke-level jitter that washes out
at the image: across three captions the rendered image changes over $88$--$94\%$
of its pixels by more than $0.05$, and the mean image difference ($0.101$--$0.139$)
is comparable to the mean parameter difference ($0.091$--$0.109$). At the same
time every draw still follows the caption, so the variation is in \emph{how} the
scene is drawn, not \emph{whether} it is the requested scene.

Where the variation sits is informative. For each of the $256$ stroke slots we
measure its standard deviation across the $8$ canvases, then average over slots;
the table below lists this per parameter, as mean $\pm$ std across slots.
\begin{center}
\small
\resizebox{\columnwidth}{!}{%
\begin{tabular}{lcc}
\toprule
Parameter & per-slot std across canvases & free? \\
\midrule
start point $x$ / $y$ & $0.225 \pm 0.049$ / $0.217 \pm 0.051$ & no \\
length                & $0.083 \pm 0.027$ & no \\
curvature             & $0.065 \pm 0.023$ & no \\
width                 & $0.048 \pm 0.018$ & no \\
colour $R/G/B$        & $0.127 \pm 0.041$ / $0.124 \pm 0.042$ / $0.128 \pm 0.048$ & no \\
opacity               & $0.081 \pm 0.026$ & no \\
free latent $11$--$13$ & $0.000 \pm 0.000$ & yes \\
\bottomrule
\end{tabular}}
\end{center}
The canvas mostly moves strokes' \emph{positions} and \emph{colours}, less their
shape (length/curvature/width), so different canvases re-arrange the strokes while
keeping the scene's structure. The three free latent channels do not vary at all:
the model uses them as a fixed internal code rather than as a diversity channel.

\begin{figure}[t]
\centering
\includegraphics[width=\columnwidth]{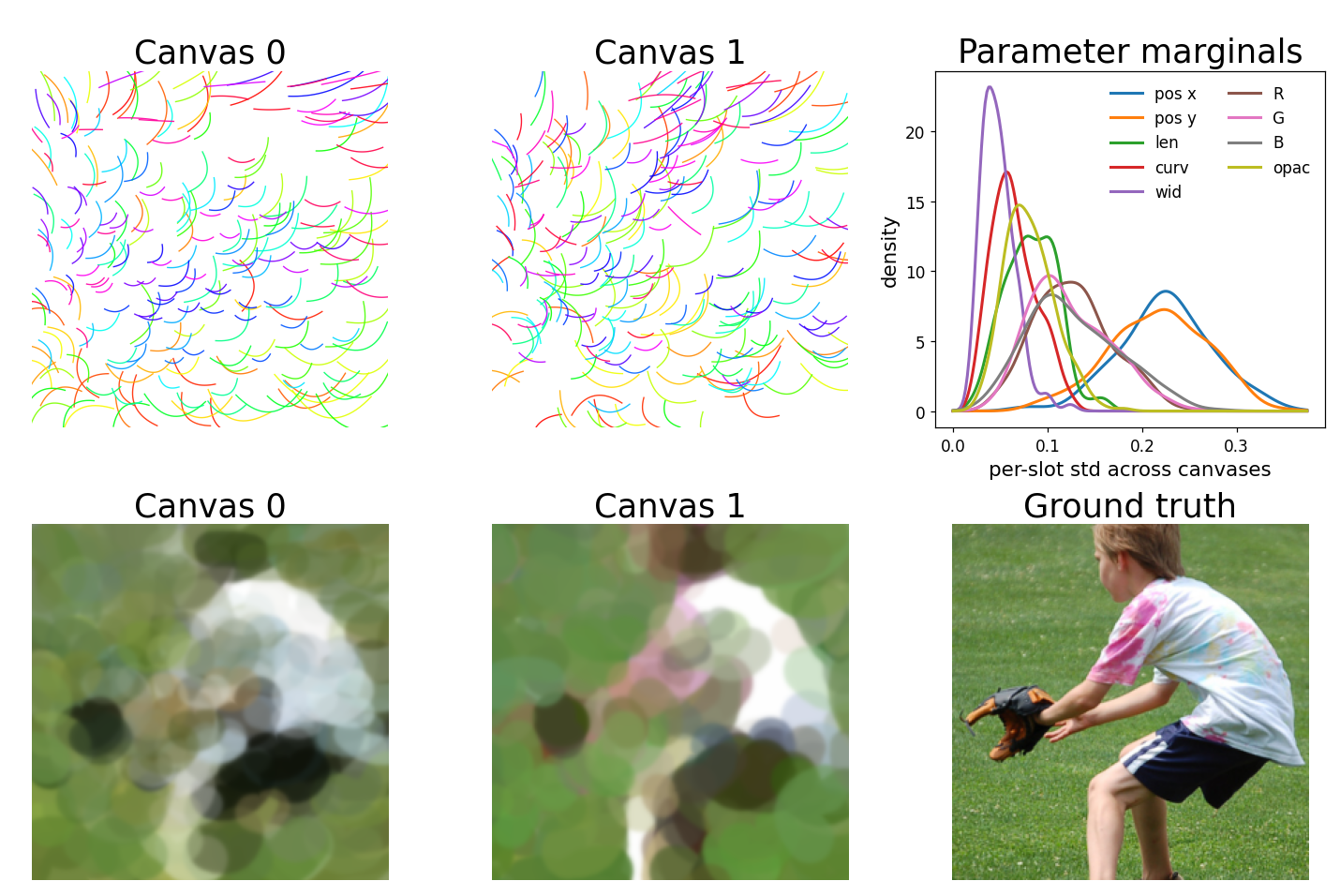}
\caption{Stroke-level view of canvas resampling on the caption ``A young boy
getting ready to catch a baseball in a grass field.'' Top row: the $256$
B\'{e}zier centre-lines for two independent draws of $z_0$, coloured by stroke
index, and the parameter marginals across $8$ canvases. Bottom row: the
canvas-0 render, the canvas-1 render, and the ground truth. No scores are shown:
the figure is about how strokes and pictures differ between draws, not about how
faithful one draw is.}
\label{fig:stroke_level}
\end{figure}

For a corpus in which each caption names a single image, this is the diversity the
objective permits: diversity that broke fidelity would leave the target, and
diversity that vanished would mean the model had collapsed to a fixed solution.
Figure~\ref{fig:stroke_level} shows the same caption drawn from two canvases plus
the ground truth: the $256$ curves differ from draw to draw, the rendered pictures
differ with them, and both remain the requested scene.

\section{Prompt generalization on SD gen}
\label{sec:suppl_sdgen}

This section reports the boundary test promised in the main paper
(Sec.~4.6). Both arms are trained on SD gen and scored on it, so this
is not a zero-shot test: it asks how each representation behaves on a corpus away
from COCO, where every caption has exactly one image. Because the pair is exact, a
model that merely recalls cannot exploit the shortcut of averaging several images
per caption.

\begin{table}[h]
\centering
\small
\begin{tabular}{lcc}
\toprule
Method & PSNR & Baseline \\
\midrule
Pictura (stroke DiT, 1 step)      & $18.8$ & $13.1$ \\
Pixel regression (DiT, 1 step)    & $24.18$ & $13.1$ \\
\bottomrule
\end{tabular}
\caption{Generation quality on SD gen against the constant-mean baseline. Both
models clear the baseline by a wide margin. Note what this does and does not show:
on a one-to-one corpus, fitting the caption-to-image map closely is compatible
with recall, so the margin establishes that both models learned the map, not that
either generalizes. The stroke generator does so with a one-step,
$3{,}584$-dimensional output.}
\label{tab:sdgen}
\end{table}

Pictura reaches $18.8$\,dB, above the $13.1$\,dB baseline. This is a boundary
check, not a claim of open-vocabulary generalization: the corpus is small and
templated, and the gap below its COCO score reflects the harder setting.

The pixel-regression row is the important one, and it is not in our favour. On
COCO the pixel arms top the table through recall; on SD gen, where every caption
has exactly one image and memorizing the caption still reproduces that image,
the $256\times256$ pixel regression reaches $24.18$\,dB---well above Pictura's
$18.8$. The honest reading is that the pixel advantage survives the move to a
prompt distribution the model was fit to, at least at this scale ($1{,}000$
images, $400$ epochs). What the stroke representation buys is the one-step,
low-dimensional output and the properties in the main paper; it does not buy a
fidelity advantage when a pixel head is trained directly on the same corpus.

\section{Inference efficiency}
\label{sec:efficiency}

Beyond quality, the two paradigms differ sharply in inference cost. We time
single-image generation with randomly initialised weights at each architecture's
real hyper-parameters, averaging $10$ images over $5$ repetitions after two warm-up
calls, and report the network forward alone and the full cost including rendering
to $256\times256$ (Table~4). Timing the architecture rather than a
trained checkpoint is what matters here: the forward-pass cost is set by the
network shape, not by the values of its weights.

Against the standard diffusion pipeline Pictura is cheaper at both stages. The
network forward costs $6.3$\,ms against $72.2$\,ms for one full-resolution DDIM
step, an $11\times$ reduction in the forward pass itself. End-to-end, including
rendering, Pictura produces a $256\times256$ image in $37.1$\,ms against
$3604$\,ms for the $50$-step model---$97\times$ faster---and $23\times$ faster than
the $35\times35$ model ($838$\,ms), which needs the same number of steps at lower
resolution. Compositing the $256$ strokes costs $30$\,ms, which is the one stage
Pictura pays and the pixel models do not; it is still well below a single DDIM
step, so it does not change the ordering.

Pictura's forward pass is exactly what the $256\times256$ DiT pixel regression
costs: both are one forward through the same trunk, and the stroke head is not
measurably more expensive than a pixel head. Against a one-step pixel regression
the margin on the network is therefore zero and end-to-end is negative. What the
paradigm buys relative to diffusion is a cheaper forward \emph{and} no sampling
loop, at the price of a bounded $30$\,ms render.

\section{The drifting-field comparison in full}
\label{sec:suppl_drift}

The remaining three arms are not scored by KID. The $35\times35$ diffusion binds
the condition but peaks early and degrades, the $256\times256$ diffusion does not
bind it, and the pixel regression memorizes, so a distribution-matching score would
not be interpretable for any of them.

This section gives the protocol and the per-arm trajectories behind the main-text
table on distribution matching (Sec.~4.2).

\paragraph{Protocol.} A drifting field~\cite{deng2025drifting} trains a generator by
pulling generated and real samples together under a repulsive term, so it needs no
discriminator and no sampling loop: like Pictura it is one-step. Because the field
is a distance, it can be computed in any embedding space, and the point of the
experiment is that this choice dominates. The three drifting arms share the same
multi100 corpus ($100$ captions $\times$ $10$ images, the last $20$ captions held
out), the same $68$M DiT backbone, and $1{,}000$ epochs; the diffusion row is the
same parameter-matched U-Net ($74.63$M) used in the main experiments, trained on the
same corpus and shown for reference. We report the kernel inception distance
(KID) on the held-out captions, computed with standard Inception pool3 features;
FID is not usable because $100$ captions cannot support a stable covariance
estimate. KID is also reported against the raw photographs
(\texttt{KIDraw}), which removes the shared renderer from the reference and
therefore rules out the possibility that the stroke result is an artefact of both
sides passing through the same renderer.

\paragraph{What each arm does over training.} Every arm reaches its best held-out
KID well before the end, and every arm then degrades. In stroke-parameter space the
held-out KID falls to $0.218$ at epoch $200$ and ends at $0.264$; the same run
scored against raw photographs gives $0.233$ at the peak, a change of less than
$0.02$. Raw-pixel space bottoms at $0.299$ and ends at $0.308$. ResNet-feature
space---the space used in~\cite{deng2025drifting}---reaches $0.377$ and ends at
$0.420$. In every arm the training loss keeps falling while the held-out KID rises
after the peak.

\paragraph{Comparing arms at a single checkpoint.} Because every arm peaks at a
different epoch, a table that mixes each arm's best value with every arm's final
value is not a like-for-like comparison: it would credit the stroke arm with
epoch $200$ and the feature arm with epoch $600$. The table therefore
fixes the criterion at the final epoch, where all three arms are complete
($0.264$ / $0.308$ / $0.420$), and reports a $95\%$ bootstrap interval over the
$20$ held-out captions for each. We generate $10$ samples per held-out caption so
that the interval reflects caption-level variability rather than single-draw
noise; the stroke interval $[0.251,0.292]$ and the pixel interval
$[0.293,0.334]$ are adjacent rather than well-separated. We verified the ordering
at four generation seeds: the point estimates move by under $0.002$, so the
ordering does not rest on a single draw.

\paragraph{Why the trajectory matters more than the ordering.} The diffusion arm on
the same corpus reaches its best held-out KID early and then degrades sharply by
the final epoch, while its training loss keeps falling. The model optimizes its own
objective more successfully as its output distribution becomes less like the data.
This is memorization, and it is the reason we report the peak and the final value
for every arm rather than a single number: on $1{,}000$ images the fastest way to
drive a sample-level objective toward zero is to reproduce individual training
targets, and the held-out captions show it first. The asymmetry between arms is in
how quickly they get there: a $50$-step diffusion can memorize pairs one at a time,
whereas a one-step distribution matcher must cover the whole conditional
distribution in a single move.

\begin{table}[h]
\centering
\small
\resizebox{\columnwidth}{!}{%
\begin{tabular}{lcccc}
\toprule
Arm & best KID & epoch & final KID & 95\% CI (final) \\
\midrule
Stroke parameters          & $\mathbf{0.218}$ & $200$ & $\mathbf{0.264}$ & $[0.251, 0.292]$ \\
Stroke params.\ (raw ref.) & $0.233$ & ---   & ---               & --- \\
Raw pixels                 & $0.299$ & $999$ & $0.308$           & $[0.293, 0.334]$ \\
ResNet features            & $0.377$ & $600$ & $0.420$           & $[0.394, 0.451]$ \\
\bottomrule
\end{tabular}}
\caption{Peak and final held-out KID per drifting arm. The table
compares arms on the final value; this table additionally shows where each arm
peaked, which is the evidence for the degradation discussed above. ``Raw ref.''
scores the stroke arm against the original photographs instead of the rendered
targets. The intervals are $95\%$ bootstraps resampling the $20$ held-out
captions, with $10$ generated samples per caption; the stroke and pixel intervals
are adjacent rather than well-separated, and the ordering holds across four
generation seeds.}
\label{tab:suppl_drift}
\end{table}

\section{Why 256 strokes}
\label{sec:suppl_strokes}

\begin{figure}[h]
\centering
\includegraphics[width=0.8\columnwidth]{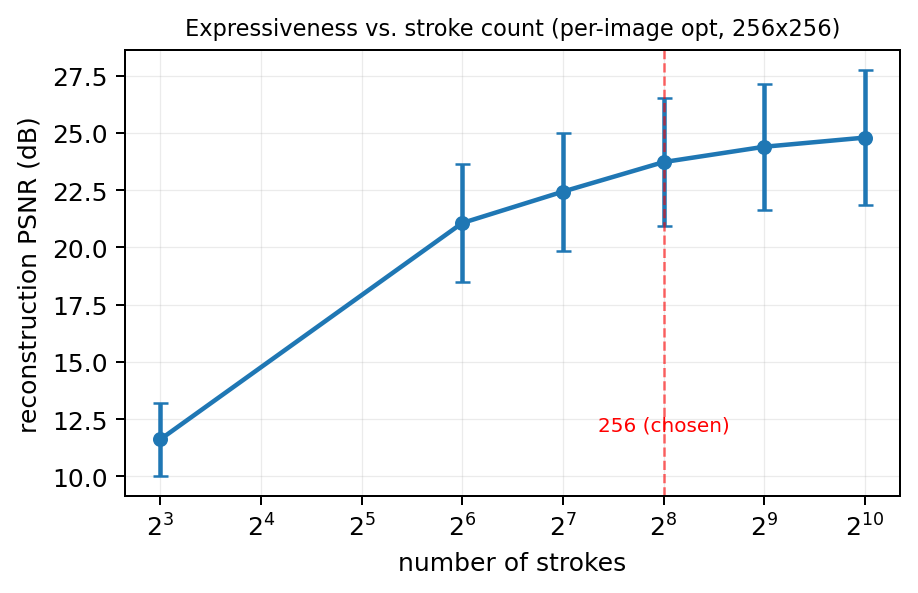}
\caption{Reconstruction PSNR vs.\ stroke count (log scale). Diminishing returns
flatten by $256$ strokes; beyond it, doubling strokes adds less than the
per-image std.}
\label{fig:stroke_count}
\end{figure}

We choose $256$ strokes by measuring how reconstruction fidelity grows with the
stroke budget. For each of $10$ images we run per-image optimization ($100$ Adam
steps) of the sdlc strokes at counts $\{8, 64, 128, 256, 512, 1024\}$ and record
the PSNR against the $256\times256$ target (Table~\ref{tab:stroke_count},
Fig.~\ref{fig:stroke_count}). The marginal return collapses quickly: doubling from
$64$ to $128$ adds $1.37$\,dB, but from $512$ to $1024$ only $0.40$\,dB---well
within the per-image std ($\approx2.9$\,dB)---so additional strokes beyond $256$
give no reliable gain. $256$ is where the curve flattens, and we adopt it as the
operating point.

\begin{table}[h]
\centering
\small
\begin{tabular}{rcc}
\toprule
Strokes & PSNR (dB) & $\Delta$ vs.\ previous \\
\midrule
8     & $11.61 \pm 1.60$ & --- \\
64    & $21.06 \pm 2.57$ & $+9.45$ \\
128   & $22.43 \pm 2.58$ & $+1.37$ \\
\textbf{256} & $\mathbf{23.72 \pm 2.81}$ & $+1.28$ \\
512   & $24.39 \pm 2.76$ & $+0.68$ \\
1024  & $24.79 \pm 2.94$ & $+0.40$ \\
\bottomrule
\end{tabular}
\caption{Reconstruction PSNR vs.\ stroke count (per-image optimization, $100$
iterations, $10$ images, $256\times256$).}
\label{tab:stroke_count}
\end{table}

\section{Editing the stroke output}
\label{sec:suppl_edit}

\begin{figure}[h]
\centering
\includegraphics[width=0.75\columnwidth]{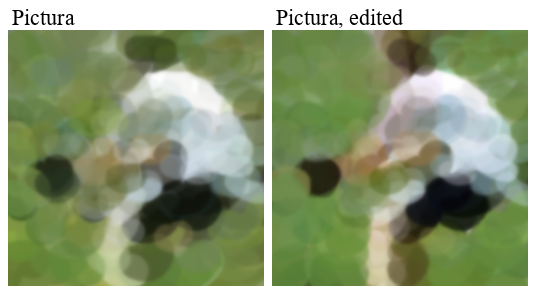}
\caption{Refining Pictura's output in stroke space. Left: Pictura's output.
Right: the same output after a few post-hoc iterations on its stroke parameters
($19.2$ to $20.1$\,dB). This refinement moves the parameter set as a whole; the
targeted, stroke-by-stroke edit is the one in Fig.~\ref{fig:editor}.}
\label{fig:edit}
\end{figure}

\begin{figure*}[t]
\centering
\includegraphics[width=\textwidth]{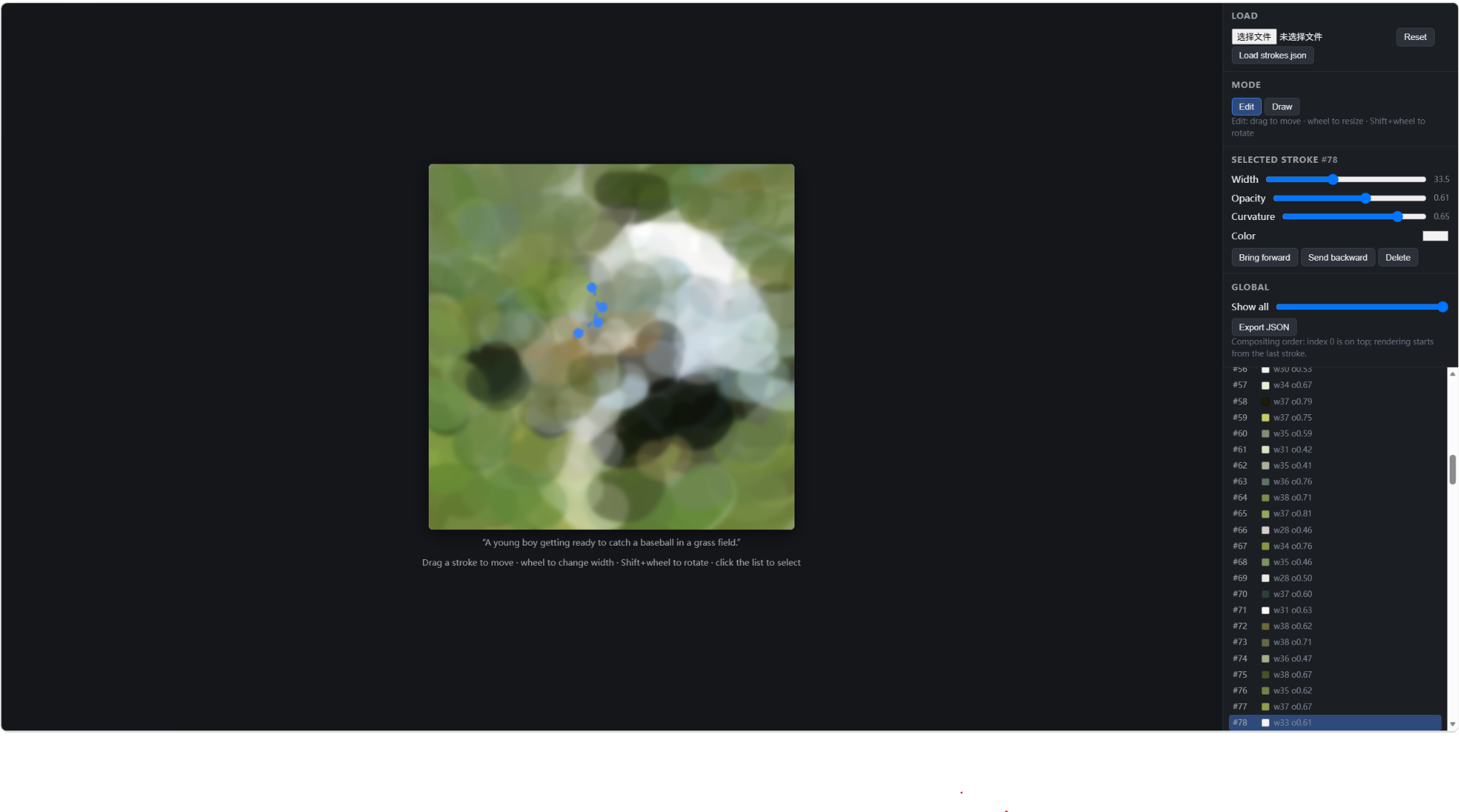}
\caption{Editing at the stroke level. The prototype loads a generated stroke set,
exposes the selected stroke's width, opacity and curvature (and its colour), and
lists all $256$ strokes with their parameters and compositing order; dragging the
handles moves the curve. An edit therefore changes the parameters a rendered stroke
is built from---position, length, curvature, width, colour, opacity---rather than
repainting pixels.}
\label{fig:editor}
\end{figure*}

Because Pictura emits $256$ parameterized curves rather than a pixel raster, its
output can be refined after generation by optimizing the stroke parameters
directly. On the baseball image a few iterations raise fidelity from $19.2$ to
$20.1$\,dB (Fig.~\ref{fig:edit}). The same parameter set can also be edited by hand:
Fig.~\ref{fig:editor} shows a prototype in which a stroke is selected, dragged and
reshaped through its width, opacity and curvature, with the stroke list exposing the
compositing order. A short screen recording of this prototype accompanies the
supplementary material (\texttt{editable\_strokes\_demo.mp4}).

\section{Why the dimensionality argument is not enough}
\label{sec:suppl_dimargument}

The main paper argues that the output representation governs whether a small-data
model binds its condition, and that dimension and structure are separable factors.
The extreme case of the dimension trend shows why dimension alone cannot carry the
explanation. Shrinking a pixel output does raise its PSNR, but pixel regression,
the limit of that trend, ignores its noise almost entirely (Table~6): its noise difference is $0.002$ against a caption difference of $0.335$. PSNR
rewards whatever lowers per-pixel error on the training prompts, and at $1{,}000$
images recall does so most cheaply. The stroke generator's $19$\,dB is therefore
not an admission of weakness against the regression arm's higher score: the two
numbers measure different things, and only the one that survives the noise probe is
evidence about generation.

\begin{table}[tb]
\centering
\small
\caption{The reachable cells of the dimension-by-structure comparison of
Sec.~4.3 in the main paper. Dimension is output size ($196{,}608$ vs $\sim3.5$k);
structure is whether the output is stroke parameters a differentiable renderer turns
into an image, or a raster emitted directly. Not all four cells are reachable---a
stroke costs $14$ values and a pixel $3$, so the stroke output cannot reach
$196{,}608$---hence the table lists only the cells we can build. Every row is a
separate model trained under the controlled setup of the main paper; at low dimension
the structured output leads pixel \emph{diffusion} by $2.7$\,dB in one step, while
pixel \emph{regression} scores higher but is excluded as generation (noise
difference $0.002$).}
\label{tab:dimstructure}
\vspace{2pt}
\resizebox{\columnwidth}{!}{%
\begin{tabular}{llcc}
\toprule
Dimension & Structure & Method & PSNR \\
\midrule
High ($196{,}608$) & no  & pixel diffusion ($50$ steps)   & $12.01$ \\
Low ($\sim3.5$k)   & no  & pixel diffusion ($50$ steps)   & $16.40$ \\
Low ($\sim3.5$k)   & yes & Pictura ($1$ step)             & $\mathbf{19.13}$ \\
\midrule
Low ($\sim3.5$k)   & no  & pixel regression (memorizes)   & $21.88$ \\
\bottomrule
\end{tabular}}

\end{table}

\section{Perceptual and semantic metrics}
\label{sec:suppl_metrics}

PSNR is a pixel-reconstruction measure, so we also score every model with three
metrics that look elsewhere: LPIPS~\cite{zhang2018unreasonable} for perceptual
distance to the paired target, CLIPScore~\cite{hessel2021clipscore} for image--text
alignment, and FID~\cite{heusel2017gans} over CLIP image features. All numbers use the same $50$ prompts and the same
ground-truth images as Table~2.

\begin{table}[h]
\centering
\small
\setlength{\tabcolsep}{5pt}
\begin{tabular}{lccc}
\toprule
Model & LPIPS $\downarrow$ & CLIPScore $\uparrow$ & FID $\downarrow$ \\
\midrule
Pictura (1 step)      & $\mathbf{0.646}$ & $0.2150$ & $\mathbf{0.73}$ \\
Diffusion $35\times35$ & $0.836$ & $0.2148$ & $0.76$ \\
Diffusion $256\times256$ & $0.658$ & $0.2147$ & $1.31$ \\
\midrule
Ground truth          & --- & $0.2143$ & --- \\
\bottomrule
\end{tabular}
\caption{Perceptual and semantic metrics on the shared $50$ prompts.}
\label{tab:metrics}
\end{table}

The three metrics do not agree, and the disagreement is informative. LPIPS places
Pictura lowest ($0.646$), full-resolution diffusion close behind ($0.658$), and the
$35\times35$ diffusion clearly worst ($0.836$). This is \emph{not} the PSNR order:
PSNR ranks the $35\times35$ diffusion second ($16.40$) and full-resolution
diffusion last ($12.01$). We therefore make no perceptual claim for the
$35\times35$ arm---it recovers on PSNR but its output is perceptually the poorest,
consistent with upsampling a $35\times35$ raster to $256$ (Sec.~D). Against the
full-resolution diffusion, Pictura is ahead on PSNR ($19.13$ against $12.01$) and
also on LPIPS ($0.646$ against $0.658$), but the LPIPS margin is smaller than the
per-image spread; we therefore lean on the PSNR comparison and read LPIPS as
consistent with it rather than as independent evidence. FID keeps the PSNR order
with a wider margin, but at $50$ images per set it is far below the sample size at
which FID is normally quoted and we report it for completeness only. CLIPScore does
not separate the models at all: all three sit within $0.0003$, and the ground-truth
images score \emph{lower} ($0.2143$) than every generated set, so on this
stereotyped-caption corpus it is uninformative and we use it in neither direction.

\section{A compositional-probe corpus}
\label{sec:suppl_realgen}

\paragraph{How the corpus was built.} Counting how often a model reproduces one
memorized caption says nothing about whether it uses the \emph{words}. To test
attribution we needed captions that differ in exactly one slot, so we built a
probe corpus rather than sampling photographs. Every caption is instantiated from
a fixed template,
\begin{center}
\small\emph{[article] [colour] [object] [relation] [article] [support]},
\end{center}
e.g.\ \emph{a red ball above a nightstand}, filled from $12$ colour words,
$43$ object nouns, $9$ spatial relations (on, above, below, under, behind, next
to, beside, near, in front of) and $43$ support nouns. We drew $1{,}000$
non-repeating combinations and rendered each with a text-to-image model
(Stable Diffusion), producing $1{,}000$ $256\times256$ images with a caption
that is symbolic rather than photographic. Captions run $6$--$10$ words, median
$7$; the least frequent colour appears $63$ times and the most frequent $100$
times, so no single colour dominates. Because the slot structure is known, any
caption can be edited at one slot while the rest is held fixed, which is what
the colour substitution exploits.

\paragraph{Why only two models.} We do not train the $256\times256$ diffusion
on this corpus, because on the photographic corpus its images respond to the
prompt but are pixel-wise far from the target and remain below the
constant-mean baseline (Sec.~A). Running a colour substitution on images whose
content is already too noisy to localise the object has no interpretation, so the
probe is restricted to the two representations that depict their captions clearly:
Pictura and the dimension-matched $35\times35$ diffusion. The exclusion follows from the finding in the main paper rather than from a choice of convenience, and among the arms that do depict their captions we keep the dimension-matched one, the stronger of the two.

\paragraph{Training.} Both models see the same $1{,}000$ image--caption pairs.
Pictura is the same $67.8$M DiT used elsewhere ($16$ blocks, width
$512$), trained for $1{,}500$ epochs; the dimension-matched pixel diffusion is
the $35\times35$ U-Net, trained for $1{,}500$ epochs. The constant-mean baseline on
this corpus is $13.50$\,dB.

\begin{figure}[t]
\centering
\includegraphics[width=\columnwidth]{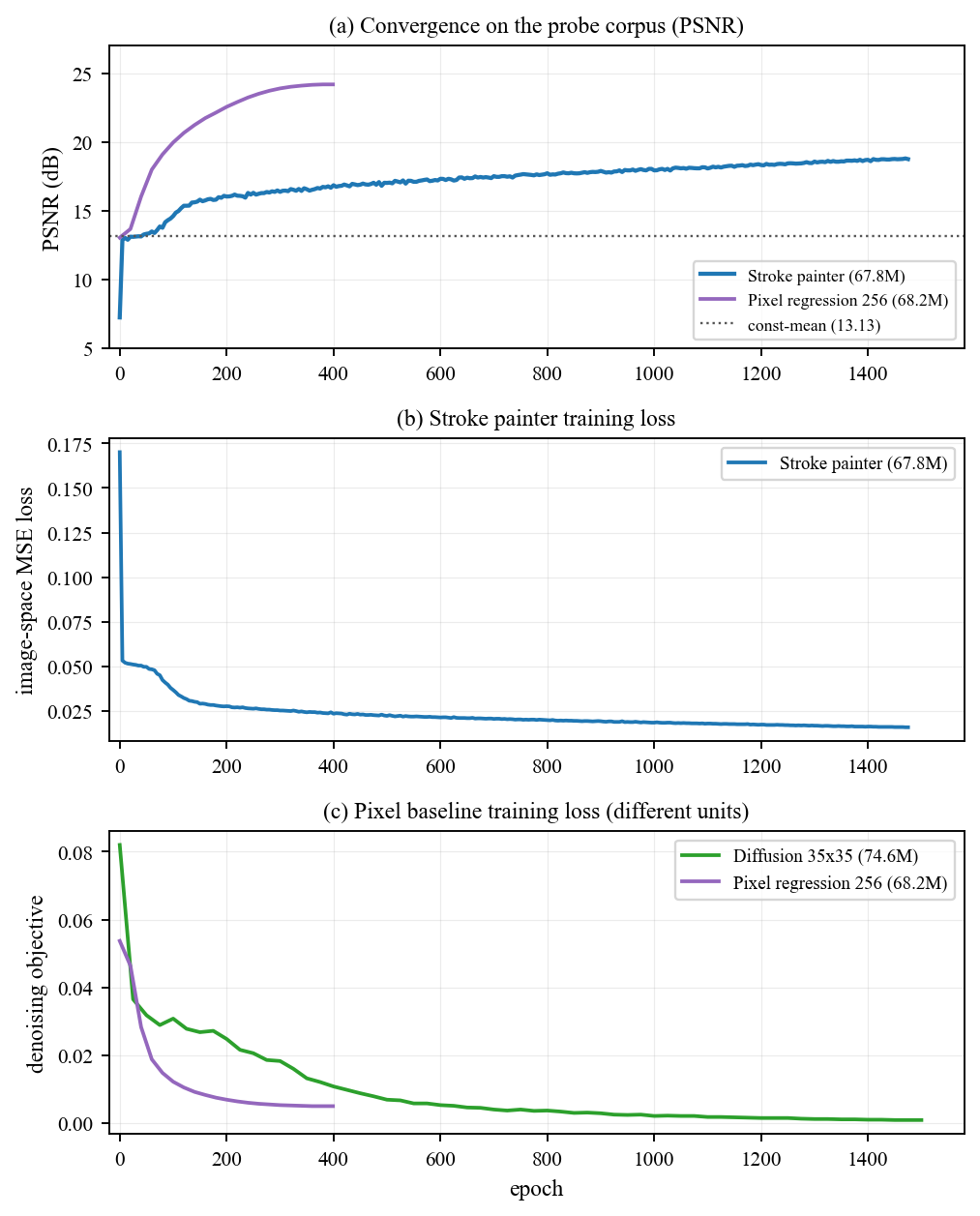}
\caption{Probe-corpus training. (a) PSNR of the two models with an image-space
MSE objective---Pictura ($7.24\to18.80$\,dB) and the $256\times256$ pixel
regression ($24.18$\,dB); dotted line is the constant-mean baseline
($13.50$). The $35\times35$ diffusion is omitted here because it optimizes a
denoising objective, so its PSNR is not on the same footing. (b) Pictura
image-space MSE loss. (c) Training losses of the two pixel baselines---the
diffusion's denoising objective and the pixel regression's image-space MSE. The
in-training PSNRs plotted
here are single-seed probes over a small prompt subset; the final scores, scored
on a fixed $200$-image subset, are in Table~\ref{tab:realgen} and are somewhat
higher. The loss panels cover different objectives and units and are therefore
kept on separate axes.}
\label{fig:realgen_curve}
\end{figure}

\paragraph{Results.} Table~\ref{tab:realgen} scores both models on a $200$-image
subset of the corpus using $50$ DDIM steps for the diffusion model and one
forward pass for Pictura. Pictura gains $+5.63$\,dB over the constant
baseline and the dimension-matched diffusion gains $+4.06$\,dB, so on a corpus
whose captions are literal descriptions of their images both representations
learn the conditional mapping. This is the point of the probe set: it is the
setting in which the colour words \emph{can} be read out, which makes the
substitution experiment below meaningful rather than a test of a model that
draws nothing at all. It is a different corpus from the photographic benchmark,
so the absolute ordering of the two representations should be read within this
table rather than against Table~2.

\begin{table}[h]
\centering
\footnotesize
\setlength{\tabcolsep}{3pt}
\begin{tabular}{lcccc}
\toprule
Model & Dim & Params & PSNR & vs.\ const. \\
\midrule
Constant-mean baseline & $196{,}608$ & -- & $13.50$ & $0.00$ \\
Pictura                 & $3{,}584$   & $67.8$M & $19.13 \pm 2.46$ & $+5.63$ \\
Diffusion $35\times35$  & $3{,}675$   & $74.6$M & $17.56 \pm 3.03$ & $+4.06$ \\
\bottomrule
\end{tabular}
\caption{Probe corpus, $200$-image subset, $50$ DDIM steps for the diffusion
model and one forward pass for Pictura. Both models are trained on all
$1{,}000$ images, so these numbers measure recall, not generalization. On this
synthetic corpus both representations learn the conditional mapping, and the
Pictura remains $1.57$\,dB above the dimension-matched pixel model.}
\label{tab:realgen}
\end{table}

\begin{figure*}[h]
\centering
\includegraphics[width=\textwidth]{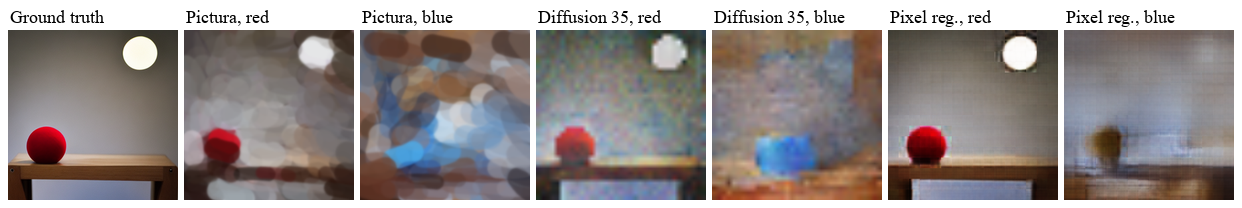}
\caption{Substituting one word in a memorized caption. Left to right: ground
truth for \emph{a red ball above a nightstand}, Pictura on that caption,
Pictura on \emph{a blue ball above a nightstand}, the $35\times35$ diffusion
on the same two captions (upsampled), and the $256\times256$ pixel regression on
the same two captions. All three representations respond to the substituted colour
word; none of them keeps the original layout while changing only the ball.}
\label{fig:colorswap}
\end{figure*}

\paragraph{Colour substitution.} Fig.~\ref{fig:colorswap} takes one memorized
caption, \emph{a red ball above a nightstand}, and substitutes a single word:
\emph{a blue ball above a nightstand}. Pictura scores $22.69$\,dB on the
original caption and drops to $13.19$\,dB on the substituted one. The colour does visibly change, but
the scene does not stay put: the ball leaves its position above the nightstand
and the surrounding strokes are redrawn. That is what an edit at the parameter level does: the substituted word changes one stroke's colour, while the other strokes, composed jointly with it, stay where they were. We claim the former, not object-level locality of the edit (Sec.~4.6). Its distance to the model's own
rendering of a genuinely different blue-ball caption, \emph{a blue ball next to
a bed}, is $122.7/255$; its distance to the original red-ball rendering is
$132.8/255$. The substituted frame is thus slightly closer to a blue-ball scene
than to the red one it was edited from: the colour word is carried, but not in
isolation---it arrives attached to a different arrangement of strokes. The $35\times35$ diffusion shows the same
dissociation: its red and blue renderings differ by $35.4/255$. The
$256\times256$ pixel-regression arm, trained on the same SD gen corpus, behaves
differently: it fits the caption-to-image map closely, its red-ball output sits
$4.3/255$ from the ground truth, and swapping the caption to \emph{blue} moves its
output by $33.6/255$---as much change as the stroke model makes. What it cannot do is hold the
scene while changing one attribute: like the other two arms, the substituted frame
is a newly composed image rather than an edit of the original layout. This is a
data-scale effect, not a
property of one representation: every model here is trained on $1{,}000$
examples and none of them factors a caption into recombinable parts.

\section{Qualitative comparison across models}
\label{sec:suppl_qual}

\begin{figure*}[h]
\centering
\includegraphics[width=\textwidth,keepaspectratio]{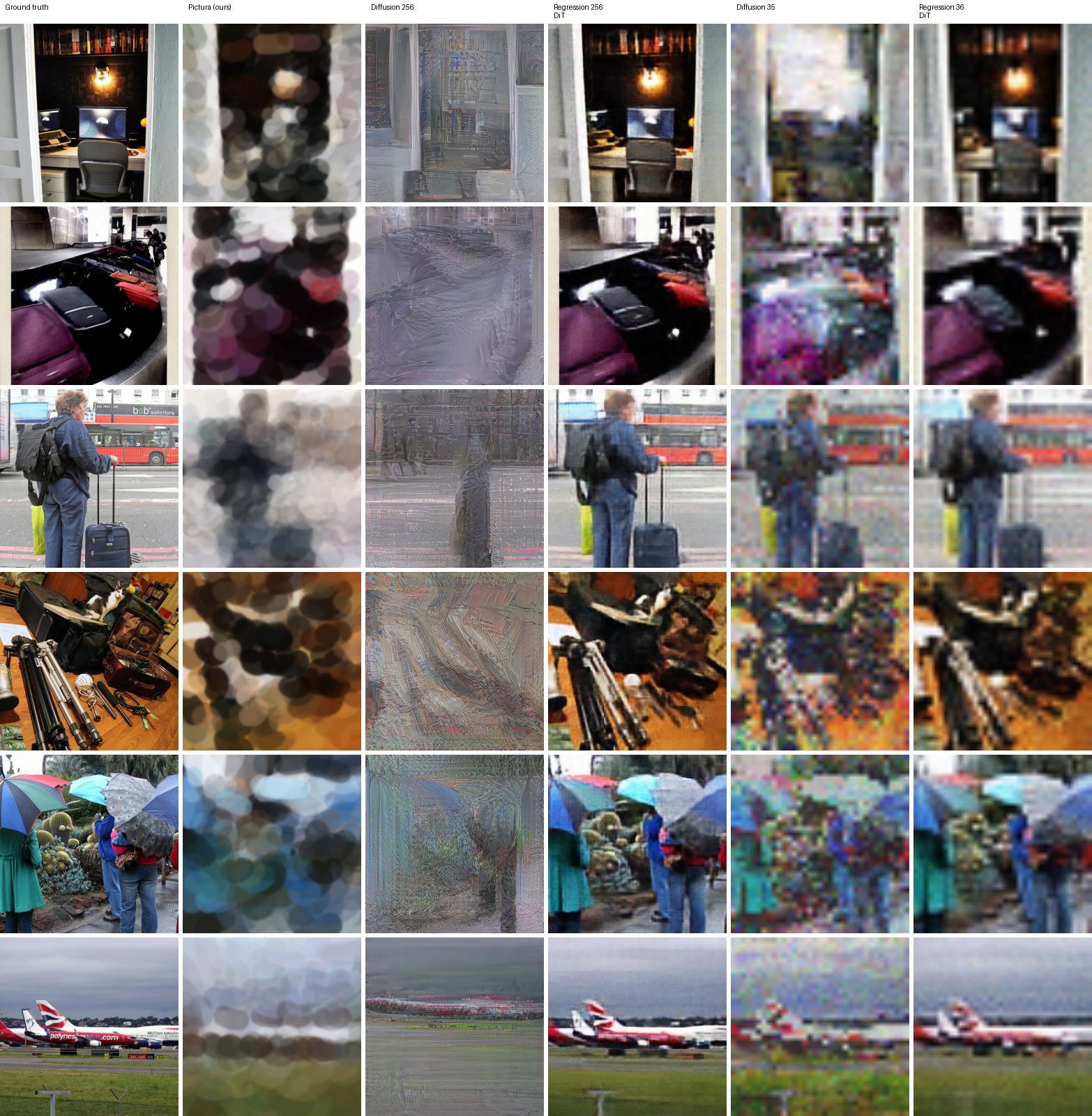}
\caption{Columns, left to right: ground truth, Pictura, standard-recipe diffusion
($256\times256$), pixel regression at $256\times256$, dimension-matched diffusion
($35\times35$, output dim $3{,}675\approx$ Pictura's $3{,}584$; shown upsampled to
$256$), and pixel regression at $36\times36$ ($3{,}888$, upsampled). Both pixel
regressions share Pictura's DiT trunk, text feature and noise canvas and differ
only in the output head, so the columns isolate the output representation. The
$256\times256$ diffusion produces a different scene per row but is pixel-wise far
from the target; the $35\times35$ column varies with the prompt but at very low
resolution, while Pictura follows each caption at full resolution.}
\label{fig:qualitative}
\end{figure*}

Figure~\ref{fig:qualitative} shows a six-column comparison on six training
prompts: ground truth, Pictura, the standard-recipe diffusion at
$256\times256$, pixel regression at $256\times256$, the dimension-matched
diffusion at $35\times35$, and pixel regression at $36\times36$. The
$256\times256$ diffusion does respond to the prompt---each row shows a different
scene---but its content is noisy and far from the target pixel-wise, which is what
its low PSNR reflects. The $35\times35$ column varies with the prompt but at very
low resolution, while Pictura follows each caption at full resolution.

The two pixel regressions are the sharpest contrast in the figure. They share
Pictura's trunk, its text features and its noise canvas, and they score higher on
this figure as well ($24.1$ and $20.5$\,dB over these six images, against $17.7$
for Pictura; Table~2 reports the $50$-prompt values, $25.20$ and
$21.88$), with an image standard deviation matching the ground truth's
($0.265$ vs.\ $0.267$). They are not blurry averages: each row is a
different scene, and each column matches the ground truth of its own prompt far
better than it matches the other five ($+15.9$ and $+12.0$\,dB). What they do not
do is respond to their noise, which is the finding of Sec.~4.4 and
of the colour-substitution test in Fig.~\ref{fig:colorswap}: the caption fixes the
image and the noise latent contributes almost nothing. The figure cannot show that
failure---a fixed noise canvas produces a perfectly plausible picture---which is
precisely why the probe is needed.

A U-Net-trunked pixel regression is degenerate (spatial std $0.021$, the same flat
field for every prompt, $10.8$\,dB) and is not shown: it would misrepresent what a
pixel regression can do. The DiT regressions are the fair comparison and are the
ones the main table reports.

\section{One-step generation across representations}
\label{sec:suppl_onestep}

\begin{figure*}[h]
\centering
\includegraphics[width=\textwidth,keepaspectratio]{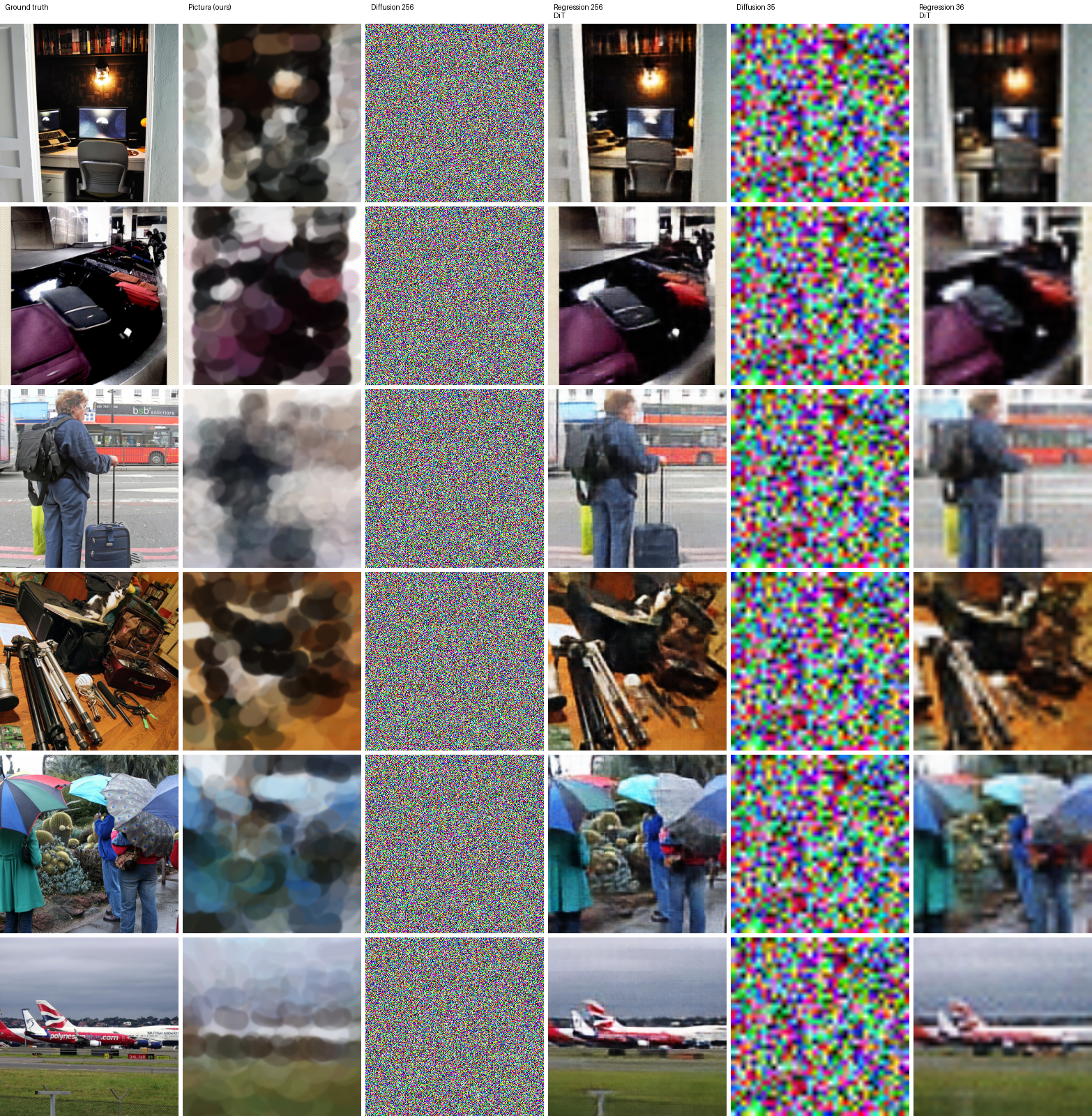}
\caption{Same six prompts and column order as Fig.~\ref{fig:qualitative}, but
every diffusion model is limited to a single forward pass: ground truth, Pictura
(one forward), pixel diffusion $256\times256$ (one DDIM step), pixel regression
$256\times256$ (one forward), pixel diffusion $35\times35$ (one DDIM step,
upsampled), and pixel regression $36\times36$ (one forward, upsampled). A truncated
DDIM step yields noise for both diffusion models; Pictura and the pixel regressions
produce an image. The regressions are unaffected by the restriction because they
have no sampling loop to truncate---they were already one-step models---which is
why this figure separates the cost of sampling from the cost of the output
representation.}
\label{fig:onestep}
\end{figure*}

Fig.~\ref{fig:onestep} repeats the comparison of Fig.~\ref{fig:qualitative}
on exactly the same six prompts and in exactly the same column order, but every
model is now given a single forward pass: one forward for Pictura and for the
pixel regressions, and one DDIM step for each diffusion model. A single truncated
DDIM step does not produce a meaningful image for either diffusion model---those
columns are essentially noise---whereas Pictura produces a different scene for
each row. The pixel regressions are unaffected by this restriction because they
have no sampling loop to truncate: they were already one-step models, and the
columns here are identical to those of Fig.~\ref{fig:qualitative}.
This isolates two effects at once. Against the diffusion models, the comparison
is at equal forward-pass budget: at one step they have not learned the
conditional mapping, whereas the stroke model has. Against the pixel regressions
the step count is not a confounder at all, so the remaining difference is the
output representation.

\begin{figure}[t]
\centering
\includegraphics[width=0.78\columnwidth]{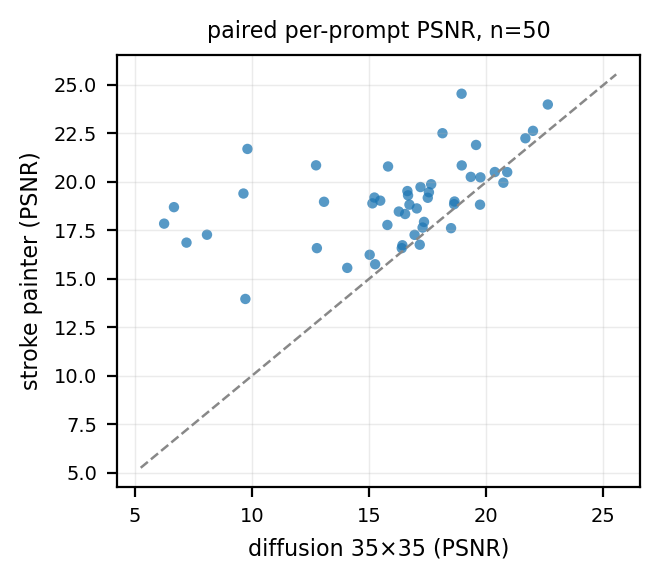}
\caption{Per-prompt PSNR, Pictura (vertical) against the $35\times35$ diffusion
(horizontal), on the shared $50$ prompts. The dashed line is $y=x$. Points above
it are prompts where Pictura is closer to the target.}
\label{fig:scatter}
\end{figure}

\paragraph{Per-prompt spread.} Fig.~\ref{fig:scatter} plots each prompt's score
under Pictura against its score under the dimension-matched diffusion, so the
paired structure behind Table~3 is visible directly. Points above
the diagonal are prompts Pictura handles better. Pictura is above the
diagonal for $45$ of $50$ prompts; the five exceptions are low-scoring prompts on
both axes, which is why a mean difference of $2.93$\,dB coexists with a per-prompt
standard deviation of roughly $3$\,dB on either side.

\section{Does upsampling inflate the $35\times35$ score?}
\label{sec:suppl_upsample}

The dimension-matched baseline renders at $35\times35$ and is upsampled to
$256\times256$ before scoring. Upsampling is a low-pass operation, and PSNR
rewards low-pass agreement, so one might worry that the reported $35\times35$
number reflects the upsampling rather than the model. We test this directly by
scoring the same $35\times35$ outputs three ways on the shared $50$ prompts.

\begin{table}[h]
\centering
\footnotesize
\setlength{\tabcolsep}{4pt}
\begin{tabular}{lccc}
\toprule
Protocol & PSNR & Baseline & Gain \\
\midrule
A.\ upsampled, $256$ GT (as reported) & $16.89$ & $12.56$ & $+4.33$ \\
B.\ upsampled, low-passed GT          & $19.14$ & $13.66$ & $+5.48$ \\
C.\ native $35\times35$               & $11.16$ & $10.43$ & $+0.73$ \\
\bottomrule
\end{tabular}
\caption{Upsampling control for the $35\times35$ baseline. A: the reported
protocol. B: the ground truth is downsampled to $35\times35$ and upsampled back,
so both sides carry the same low-pass content. C: prediction and ground truth are
compared at $35\times35$ natively. This control uses an independently drawn
$50$-image subset, so its protocol-A row ($16.89$ vs.\ $12.56$) differs from the
main table ($16.40 \pm 0.04$ vs.\ $12.51$) by sampling alone; only the ordering
and the relative size of the three protocols should be read here, not the absolute
values.}
\label{tab:upsample}
\end{table}

Protocol B is the decisive one: if the advantage came from upsampling, matching
the ground truth's bandwidth would erase it. It does not---the gain rises from
$+4.33$ to $+5.48$\,dB, because once both sides share the same bandwidth the model
is no longer penalized for detail it cannot represent. The reported number is
therefore not an upsampling artefact. Protocol C shows the complementary fact:
measured natively the same model clears its own baseline by only $+0.73$\,dB, so
most of its $35\times35$ fidelity is low-frequency structure rather than fine
detail. Both statements are consistent with the reading in the main paper that a
$35\times35$ pixel output recovers the condition along a coarse, low-frequency
channel.

\section{Scaling with data and capacity}
\label{sec:suppl_scale}

\begin{figure*}[h]
\centering
\includegraphics[width=\textwidth]{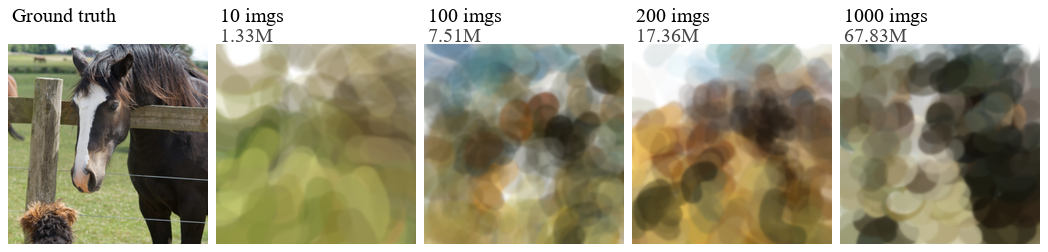}
\caption{Scale comparison on one shared prompt (``A horse looking over a fence at
a dog.''). Left to right: ground truth, and Pictura at $10$/100/200/1000 images
with $1.3$M/$7.5$M/$17.4$M/$67.8$M parameters. Structure and fidelity improve with
data and capacity.}
\label{fig:scale}
\end{figure*}

Figure~\ref{fig:scale} generates a single shared prompt (``A horse looking over a
fence at a dog.'') with four Pictura models trained on increasing data and
capacity: $10$ images/$1.3$M parameters, $100$/$7.5$M, $200$/$17.4$M, and
$1{,}000$/$67.8$M. The prompt belongs to all four nested training subsets, so each
model was trained on it. The rendered image gains structure and fidelity with
scale, consistent with the representation supporting further headroom as data and
compute grow.

\section{Held-out evaluation}
\label{sec:suppl_holdout}

Table~\ref{tab:holdout} repeats the comparison of Table~2 on $60$
images that none of the models was trained on. It is an overfitting diagnostic, not
a capability comparison: every arm trained on $1{,}000$ images loses most of its
training-set score, and the $13$\,dB spread of the main table collapses to
$0.35$\,dB. Ranking on unseen images reads the opposite way---the two strongest
arms on the training set fall the hardest---which is why we report PSNR as a
conditioning score on the training corpus and make no claim about held-out
reconstruction.

\begin{table}[t]
\centering
\small
\caption{The same models on $60$ images none of them was trained on, scored exactly
as in Table~2. Mean $\pm$ std over five training runs.}
\label{tab:holdout}
\vspace{2pt}
\resizebox{\columnwidth}{!}{%
\begin{tabular}{lcc}
\toprule
Method & PSNR (train) & PSNR (held-out) \\
\midrule
\textbf{Pictura (stroke DiT)}                 & $\mathbf{19.13 \pm 0.12}$ & $9.63 \pm 0.16$ \\
Pixel diffusion, standard ($50$ steps)        & $12.01 \pm 0.09$ & $9.52 \pm 0.20$ \\
Pixel diffusion, $35\times35$ ($50$ steps)    & $16.40 \pm 0.04$ & $9.84 \pm 0.19$ \\
Pixel regression, $36\times36$ (1 step)       & $21.88 \pm 0.01$ & $9.75 \pm 0.31$ \\
Pixel regression, $256\times256$ (1 step)     & $25.20 \pm 0.29$ & $9.49 \pm 0.24$ \\
\bottomrule
\end{tabular}}

\end{table}
\end{document}